\documentclass[lettersize,journal]{IEEEtran}
\usepackage{amsmath,amsfonts}
\usepackage{array}
\usepackage[caption=false,font=normalsize,labelfont=sf,textfont=sf]{subfig}
\usepackage{textcomp}
\usepackage{stfloats}
\usepackage{url}
\usepackage{verbatim}
\usepackage{graphicx}
\usepackage{cite}

\usepackage{xcolor}
\usepackage{balance}
\usepackage{mathtools}
\usepackage{multirow}
\usepackage{color}
\usepackage{enumitem}
\usepackage{booktabs}

\usepackage{tikz}
\usetikzlibrary{pgfplots.groupplots}
\usetikzlibrary{external}
\usepackage{pgfplots,pgfplotstable}
\pgfplotsset{compat=1.5.1}
\def\addlegendimage{\csname pgfplots@addlegendimage\endcsname}
\usetikzlibrary{patterns}
\usetikzlibrary{pgfplots.groupplots}
\usetikzlibrary{
  pgfplots.colorbrewer,
}
\pgfplotsset{
  cycle list/.define={my marks}{
    every mark/.append style= {solid,fill=\pgfkeysvalueof{/pgfplots/mark list fill}},mark=*\\
    every mark/.append style={solid,fill=\pgfkeysvalueof{/pgfplots/mark list fill}},mark=square*\\
    every mark/.append style={solid,fill=\pgfkeysvalueof{/pgfplots/mark list fill}},mark=triangle*\\
    every mark/.append style={solid,fill=\pgfkeysvalueof{/pgfplots/mark list fill}},mark=diamond*\\
  },
}
\usepgfplotslibrary{groupplots}

\usepackage[linesnumbered,ruled,vlined]{algorithm2e}

\SetCommentSty{mycommfont}

\newcommand{\m}{\textsf{CASH}}

\newcommand{\myhref}[3][black]{\href{#2}{\color{#1}{#3}}}%
\newcommand{\codelink}{https://github.com/yy-ko/cash}

\definecolor{bleudefrance}{rgb}{0.19, 0.55, 0.91}
\definecolor{mygreen2}{rgb}{0.2,0.7,0.310}
\definecolor{mygreen}{rgb}{0.0, 0.5, 0.0}
\definecolor{mygold}{rgb}{0.914,0.725,0.431}
\definecolor{mypurple}{rgb}{0.514,0.325,0.831}

\usepackage{hyperref} %<--- Load after everything else

\begin{document}

% \title{\m: Context-Aware Self-Supervised Learning for Accurate Hyperedge Prediction}
\title{Enhancing Hyperedge Prediction with Context-Aware Self-Supervised Learning}

\author{
        % Yunyong Ko$^{\textsuperscript{\orcidicon{0000-0003-1283-4697}}}$,
        Yunyong Ko,
        Hanghang Tong,~\IEEEmembership{Fellow,~IEEE,}
        % Hanghang Tong$^{\textsuperscript{\orcidicon{0000-0003-4405-3887}}}$,~\IEEEmembership{Fellow,~IEEE,}
        and~Sang-Wook Kim\IEEEauthorrefmark{1},~\IEEEmembership{Senior Member,~IEEE}
        % and~Sang-Wook Kim$^{\textsuperscript{\orcidicon{0000-0002-6345-9084}}}$\IEEEauthorrefmark{1},~\IEEEmembership{Member,~IEEE}% <-this % stops a space
\thanks{Yunyong Ko is with the School of Computer Science and Engineering, Chung-Ang University Seoul, Korea. E-mail: yyko@cau.ac.kr.}% <-this % stops a space
\thanks{Hanghang Tong is with the Department of Computer Science, University of Illinois at Urbana-Champaign, Urbana, IL, USA. E-mail: htong@illinois.edu.}% <-this % stops a space
\thanks{Sang-Wook Kim is with the Department of Computer Science, Hanyang University, Seoul, South Korea. E-mail:wook@hanyang.ac.kr.}% <-this % stops a space
\thanks{\IEEEauthorrefmark{1}Corresponding author.}
% \thanks{Manuscript received September XX, 2023.}
}


\markboth{IEEE Transactions on Knowledge and Data Engineering,~Vol.~0, No.~0,~2025}%
{Ko \MakeLowercase{\textit{et al.}}: Enhancing Hyperedge Predictiomn with Context-Aware Self-Supervised Learning}


\maketitle

\begin{abstract}
Hypergraphs can naturally model \textit{group-wise relations} (e.g., a group of users who co-purchase an item) as \textit{hyperedges}.
\emph{Hyperedge prediction} is to predict future or unobserved hyperedges, 
which is a fundamental task in many real-world applications (e.g., group recommendation). 
Despite the recent breakthrough of hyperedge prediction methods, the following challenges have been rarely studied: 
(\textbf{C1}) \emph{How to aggregate the nodes in each hyperedge candidate for accurate hyperedge prediction?} and (\textbf{C2}) \emph{How to mitigate the inherent data sparsity problem in hyperedge prediction?} 
To tackle both challenges together, in this paper, we propose a novel hyperedge prediction framework (\textbf{{\m}}) that employs (1) \emph{context-aware node aggregation} to precisely capture complex relations among nodes in each hyperedge for (C1) and (2) \emph{self-supervised contrastive learning} in the context of hyperedge prediction to enhance hypergraph representations for (C2). 
Furthermore, as for (C2), we propose a \textit{hyperedge-aware augmentation} method to fully exploit the latent semantics behind the original hypergraph and consider both node-level and group-level contrasts (i.e., \textit{dual contrasts}) for better node and hyperedge representations.
Extensive experiments on six real-world hypergraphs reveal that {\m} consistently outperforms all competing methods in terms of the accuracy in hyperedge prediction and each of the proposed strategies is effective in improving the model accuracy of {\m}.
For the detailed information of {\m}, we provide the code and datasets at: \url{\codelink}.
\end{abstract}


\IEEEpeerreviewmaketitle

\section{Introduction}\label{sec:intro}

\begin{figure}[t]
    \centering
    \includegraphics[width=0.435\textwidth]{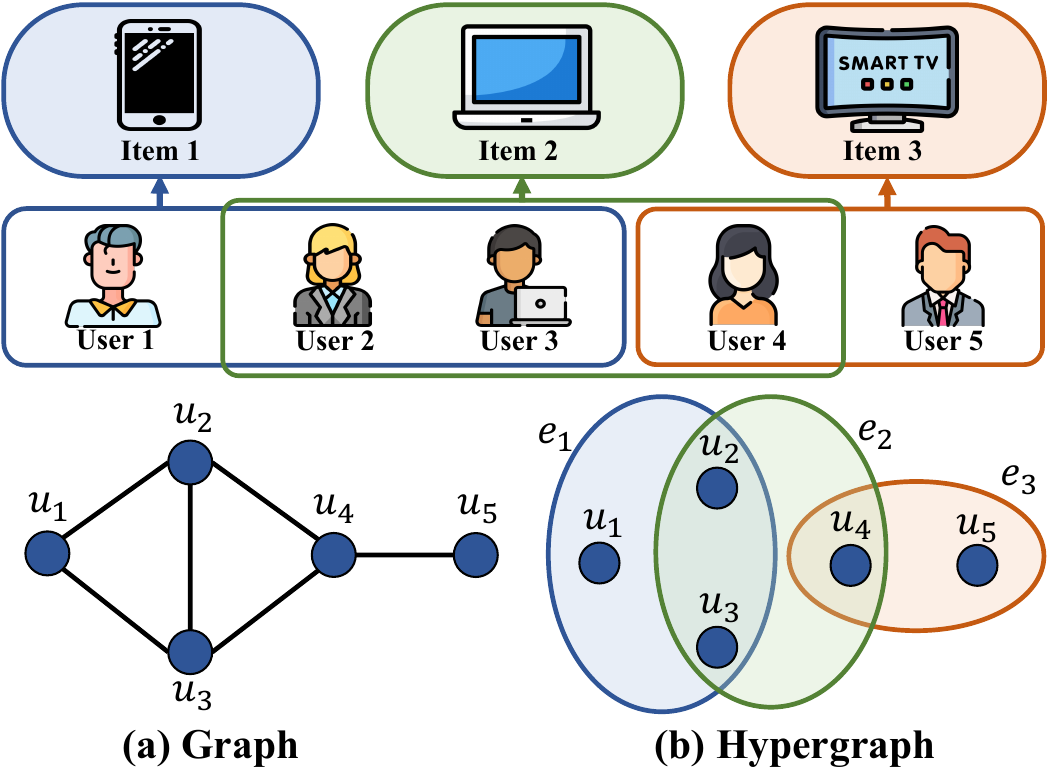}
    \caption{Group-wise relations in e-commerce networks modeled as (a) a graph and (b) a hypergraph, where each hyperedge represents an item co-purchased by a group of users.}\label{fig:intro-hypergraph}
    \vspace{-4mm}
\end{figure}

% real-worl networks
Graphs are widely used to model real-world networks, where a node represents an object and an edge does a pair-wise relation between two objects. 
In real-world networks, however, high-order relations (i.e., \textit{group-wise relations}) are prevalent~\cite{zhou2006learning,benson2018simplicial,do2020structural,amburg2020clustering,comrie2021hypergraph,choe2022midas,jiang2018hyperx,jiang2022hypergraph,sun2023self}, 
such as (1) an item co-purchased by a group of users in e-commerce networks, 
(2) a paper co-authored by a group of researchers in collaboration networks, 
and (3) a chemical reaction co-induced by a group of proteins in protein-protein interaction networks. 
Modeling such group-wise relations by an ordinary graph could lead to unexpected information loss.
As shown in Figure~\ref{fig:intro-hypergraph}(a), for example, 
the group-wise relation among users $u_1$, $u_2$, and $u_3$ for item 1 (i.e., the three users co-purchase the same item 1) is missing in a graph,
instead, there exist three separate pair-wise relations (i.e., clique).

% Hypergraph
A \textit{hypergraph}, a generalized graph structure, can naturally model such high-order relations without any information loss, 
where a group-wise relation among an arbitrary number of objects is modeled as a \textit{hyperedge}, e.g., the group-wise relation among users $u_1$, $u_2$, and $u_3$ for item 1 is modeled as a single hyperedge $e_1=\{u_1,u_2,u_3\}$ (blue ellipse) as shown in Figure~\ref{fig:intro-hypergraph}(b). 
As a special case, if the size of all hyperedges (i.e., the number of nodes in a hyperedge) is restricted to 2, a hypergraph is degeneralized to a graph.
Thanks to their powerful expressiveness, 
Hypergraph-based network learning methods~\cite{yoon2020expansion,hwang2022ahp,dong2020hnhn,yadati2020nhp} have been widely explored and have consistently outperformed graph-based methods across various downstream tasks, including node classification~\cite{feng2019hypergraph}, node ranking~\cite{chitra2019random}, link prediction~\cite{yoon2020expansion}, and anomaly detection~\cite{lee2022hashnwalk}.

% Hyperedge prediction
\textit{Hyperedge prediction} (i.e., link prediction on hypergraphs) is a fundamental task in many real-world applications in the fields of recommender systems~\cite{xia2022hypergraph,wang2020next,han2022dh,li2022enhancing,yu2021self,zhang2021double,xia2021self}, social network analysis~\cite{liben2003link,lu2011link,li2013link,zhang2018beyond}, bioinformatics~\cite{liu2017computational,vaida2019hypergraph}, and so on,
which predicts future or unobserved hyperedges based on an observed hypergraph structure.
For example, it predicts (a) an item that a group of users are likely to co-purchase in recommender systems and (b) a set of proteins that could potentially co-induce a chemical reaction in bioinformatics.
The process of hyperedge prediction is two-fold~\cite{hwang2022ahp}: 
given a hypergraph, (1) (\textbf{hypergraph encoding}) the embeddings of nodes are produced by a hypergraph encoder (e.g., hypergraph neural networks (HGNNs)~\cite{tu2018structural,yadati2019hypergcn,feng2019hypergraph,dong2020hnhn,ding2020more,yang2022semi,wu2022hypergraph,chien2021you}) and (2) (\textbf{hyperedge candidate scoring}) the embeddings of nodes in each hyperedge candidate are \textit{aggregated} and fed into a prediction model (e.g., MLP) to decide whether the candidate is real or not.

% Motivation
\vspace{1mm}
\noindent
\textbf{Challenges.}
Although many existing methods have been proposed to improve hyperedge prediction~\cite{yadati2020nhp,zhang2020hypersagnn,yoon2020expansion,nguyen2021centsmoothie,hwang2022ahp,patil2020negative}, 
the following challenges remain largely under-explored:

% Challenge 1
\textbf{(C1) Node aggregation}. ``\textit{How to aggregate the nodes in each hyperedge candidate for accurate hyperedge prediction?}" 
Intuitively, the formation of group-wise relations (i.e., hyperedges) is more complex than that of pair-wise relations (i.e., edges). 
For example, the number of nodes engaged and their influences could be different depending on hyperedges. 
On the other hand, for edges to represent pair-wise relations, the number of nodes engaged is always 2. 
Such complex and subtle properties of hyperedge formation, however, have rarely been considered in existing methods.
Instead, they simply aggregate the nodes in each hyperedge candidate by using heuristic rules~\cite{hwang2022ahp,patil2020negative} (e.g., average pooling). 
Thus, they fail to precisely capture the complex relations among nodes, which eventually results in accuracy degradation.

% Challenge 2
\textbf{(C2) Data sparsity}. ``\textit{How to mitigate the inherent data sparsity problem in hyperedge prediction?}"
% In real-world hypergraphs, there exist only a few group-wise relations, which tend to be even fewer than pair-wise relations~\cite{yu2021self,zhang2021double}.
While hypergraphs are effective in modeling complex relationships, real-world networks are inherently sparse, meaning that most objects have only a few relationships~\cite{ko2021mascot}.
The data sparsity tends to be more serious in hypergraphs than in ordinary graphs~\cite{yu2021self,zhang2021double} because the potential number of hyperedges is much greater than that of pair-wise edges (i.e., $2^{|V|} >> |V|^2$, where $|V|$ is the number of nodes).
Although existing works have studied (a) HGNNs to effectively learn the hypergraph structure based on the limited number of observed hyperedges~\cite{yadati2020nhp,zhang2020hypersagnn,yoon2020expansion,nguyen2021centsmoothie} and (b) negative samplers to select negative examples (non-existing hyperedges) useful in model training~\cite{hwang2022ahp,patil2020negative}, 
% they still suffer from the data sparsity problem.
the data sparsity remains a significant challenge, particularly for hyperedge prediction.

% examples 

% Proposed method
\vspace{1mm}
\noindent
\textbf{Our work.}
To tackle the aforementioned challenges together, we propose a novel hyperedge prediction framework, 
named \textbf{\underline{C}}ontext-\textbf{\underline{A}}ware \textbf{\underline{S}}elf-supervised learning for \textbf{\underline{H}}yperedge prediction (\textbf{{\m}}). 
{\m} employs the two key strategies: 

\textbf{(1) Context-aware node aggregation}. 
To aggregate the nodes in each hyperedge candidate while considering their complex and subtle relations among them precisely,
we propose a method of \textit{context-aware node aggregation} that calculates different degrees of influences of the nodes in a hyperedge candidate to its formation and integrates the contextual information into the node aggregation process.

\textbf{(2) Self-supervised contrastive learning}. 
To alleviate the inherent data sparsity problem, we incorporate \textit{self-supervised contrastive learning}~\cite{you2020graph,zhu2020deep,lee2022relational,lee2022m} into the training process of {\m}, 
providing complementary information to improve the accuracy of hyperedge prediction.
Specifically, we propose a method of \textit{hyperedge-aware augmentation} to generate two augmented hypergraphs that preserve the structural properties of the original hypergraph,
which enables {\m} to fully exploit the latent semantics behind the original hypergraph.
We also consider not only \textit{node-level} but also \textit{group-level} contrasts in contrastive learning to better learn node and hyperedge embeddings (i.e., \textit{dual contrastive loss}).

% Experimental results
Lastly, we conduct extensive experiments on real-world hypergraphs to evaluate {\m},
which reveal that 
(1) (\textit{Accuracy}) {\m} consistently outperforms \textit{all} competing methods in terms of the accuracy in hyperedge prediction (up to $4.78\%$ higher than the best state-of-the-art method~\cite{hwang2022ahp}), 
(2) (\textit{Effectiveness}) the proposed strategies of {\m} are \textit{all} effective in improving the accuracy of {\m}, 
(3) (\textit{Insensitivity}) {\m} could achieve high accuracy across a wide range of values of hyperparameters (i.e., low hyperparameter sensitivity), 
and (4) (\textit{Scalability}) {\m} provides (almost) linear scalability in training with the increasing size of hypergraphs.

\vspace{1mm}
\noindent
\textbf{Contributions.}
The main contributions of our work are summarized as follows.
\begin{itemize}[leftmargin=10pt]
    \item \textbf{Challenges}: We point out two important but under-explored challenges of hyperedge prediction: (\textbf{C1}) the node aggregation of a hyperedge candidate and (\textbf{C2}) the data sparsity.
    \item \textbf{Framework}: We propose a novel hyperedge prediction framework, {\m} that employs (1) a \textit{context-aware node aggregation} for (C1) and (2) \textit{self-supervised learning} equipped with \textit{hyperedge-aware augmentation} and \textit{dual contrastive loss} for (C2).
    \item \textbf{Evaluation}: Through extensive evaluation using six real-world hypergraphs, we demonstrate the superiority of {\m} in terms of (1) accuracy, (2) effectiveness, (3) insensitivity, (4) efficiency, and (5) scalability.
\end{itemize}

\noindent
For reproducibility, we provide the code and datasets used in this paper at 
\myhref{\codelink}{\codelink}

\section{Related Works}\label{sec:related}

In this section, we introduce existing hyperedge prediction methods and self-supervised hypergraph learning methods and explain their relation to our work.

\vspace{1mm}
\noindent
\textbf{Hyperedge prediction methods.}
There have been many works to study hyperedge prediction; 
they mostly formulate the hyperedge prediction task as a classification problem~\cite{tu2018structural,yoon2020expansion,yadati2020nhp,zhang2020hypersagnn,hwang2022ahp}.
Expansion~\cite{yoon2020expansion} represents a hypergraph into multiple \textit{n}-projected graphs and applies a logistic regression model to the projected graphs to predict unobserved hyperedges.
HyperSAGNN~\cite{zhang2020hypersagnn} employs self-attention-based graph neural networks for hypergraphs to learn hyperedges with variable sizes and estimates the probability of each hyperedge candidate being formed.
NHP~\cite{yadati2020nhp} adopts hyperedge-aware graph neural networks~\cite{kipf2016semi} to learn the node embeddings in a hypergraph and aggregates the learned embeddings of nodes in each hyperedge candidate via max-min pooling for hyperedge prediction.
AHP~\cite{hwang2022ahp}, a state-of-the-art hyperedge prediction method, employs an adversarial training-based model to generate negative hyperedges for use in the model training for hyperedge prediction and adopts max-min pooling as a node aggregation method.

These methods, however, suffer from the data sparsity problem since they rely only on a small number of existing group-wise relations (i.e., observed hyperedges).
On the other hand, our {\m} incorporates self-supervised contrastive learning into the context of hyperedge prediction, which provides complementary information for obtaining better node and hyperedge representations, thereby alleviating the data sparsity problem eventually.

\vspace{1mm}
\noindent
\textbf{Self-supervised hypergraph learning.}
Recently, there have been a handful of works to study self-supervised learning on hypergraphs~\cite{du2021hypergraph,xia2021self,zhang2021double,yu2021self,lee2022m}.
HyperGene~\cite{du2021hypergraph} adopts bi-level (node- and hyperedge-level) self-supervised tasks to effectively learn group-wise relations. 
However, it adopts a clique expansion to transform a hypergraph into a simple graph, which incurs a significant loss of high-order information and does not employ contrastive learning.
TriCL~\cite{lee2022m} employs tri-level (node-, group-, and membership-level) contrasts in contrastive hypergraph learning. 
This method, however, has been studied only in the \textit{node-level} task (e.g., node classification) but not in the hyperedge-level task (i.e., hyperedge prediction) that we focus on. 
Thus, TriCL does not tackle the node aggregation challenge (C1) that we point out.
Also, TriCL adopts simple random hypergraph augmentation methods~\cite{you2020graph} that do not consider the structural properties of the original hypergraph.
HyperGCL~\cite{wei2022augmentations} employs two hyperedge augmentation strategies to build contrastive views of a hypergraph. 
HyperGCL (i) directly drops random hyperedges and (ii) masks nodes in each hyperedge randomly (i.e., hyperedge membership masking).
This method, however, does not take into account the structural properties of the original hypergraph.
On the other hand, our proposed hyperedge-aware augmentation method builds two contrastive views that preserve the structural properties of the original hypergraph.

In the context of recommendations, 
DHCN~\cite{xia2021self}, a session-based recommendation method,
models items in a session as a hyperedge and captures the group-wise relation of each session by employing a group-level contrast.
However, DHCN adopts a clique expansion, incurring information loss, 
and does not consider a node-level contrast in the model training.
$\text{S}^2$-HHGR~\cite{zhang2021double} is a self-supervised learning method for group recommendation, 
which employs a hierarchical hypergraph learning method to capture the group interactions among users.
However, they do not consider a group-level contrast.
MHCN~\cite{yu2021self}, a social recommendation method, models three types of social triangle motifs as hypergraphs. 
However, the motifs used in MHCN are only applicable to a recommendation task, but not to a general hypergraph learning task such as the hyperedge prediction that we focus on in this paper.

\vspace{1mm}
\noindent
\textbf{Subgraph representation learning.}
(C1) the challenge of node aggregation in a hyperedge candidate is closely related to subgraph representation learning, 
as a hyperedge can be viewed as a small subgraph. 
We discuss subgraph representation learning within the context of hyperedge prediction.
Subgraph representation learning~\cite{alsentzer2020subgraph,wang2021glass,hamidi2022subgraph} aims to learn meaningful representations of subgraphs, rather than entire graphs, capturing localized structural information around the subgraph. 
Therefore, in the context of hyperedge prediction, 
subgraph representation learning can serve as a tool for learning representations of hyperedge candidates that can capture the localized context information within a hyperedge candidate. 
SubGNN~\cite{alsentzer2020subgraph} adopts a novel subgraph routing mechanism to capture complex topology and positional information.
GLASS~\cite{wang2021glass} (GNN with LAbeling trickS for Subgraph) employs a simple yet powerful "max-zero-one" labeling trick to distinguish nodes inside and outside subgraphs.

\section{The Proposed Method: {\m}}\label{sec:proposed}
In this section, we present a novel hyperedge prediction framework, 
\textbf{\underline{C}}ontext-\textbf{\underline{A}}ware \textbf{\underline{S}}elf-supervised learning for \textbf{\underline{H}}yperedge prediction (\textbf{{\m}}).
First, we introduce the notations and define the problem that we aim to solve (Section~\ref{sec:proposed-problem}).
Then, we describe two key strategies of {\m}: context-aware node aggregation (Section~\ref{sec:proposed-primiary}) and self-supervised learning (Section~\ref{sec:proposed-auxiliary}). 
Finally, we analyze the space and time complexity of {\m} (Section~\ref{sec:proposed-complexity}).

\subsection{Problem Definition}\label{sec:proposed-problem}

\subsubsection{\textbf{Notations}}
The notations used in this paper are described in Table~\ref{table:notations}.
Formally, a \textit{hypergraph} is defined as $H=(V,E)$, where $V=\{v_1, v_2, ..., v_{|V|}\}$ is the set of nodes and $E=\{e_1, e_2, ..., e_{|E|}\}$ is the set of hyperedges.
The node features are represented by the matrix $\mathbf{X} \in \mathbb{R}^{|V|\times F}$,
where each row $x_i$ represents the $F$-dimensional feature of node $v_i$.
Each hyperedge $e_j \in E$ contains an arbitrary number of nodes and has a positive weight $w_{jj}$ in a diagonal matrix $W \in \mathbb{R}^{|E|\times |E|}$.
A hypergraph can generally be represented by an \textit{incidence} matrix $\mathbf{H}\in \{0,1\}^{|V|\times |E|}$,
where each element $h_{ij}=1$ if $v_i \in e_j$, and $h_{ij}=0$ otherwise.
To denote the degrees of nodes and hyperedges,
we use \textit{diagonal} matrices $\mathbf{D}^V$ and $\mathbf{D}^E$, respectively.
In $\mathbf{D}^V$, each element $d^v_{ii}=\sum_{j=1}^{|E|}w_{jj}h_{ij}$ represents the sum of the weights of node $v_i$'s incident hyperedges,
and in $\mathbf{D}^E$, each element $d^e_{jj}=\sum_{i=1}^{|V|}h_{ij}$ represents the number of nodes in hyperedge $e_j$.
We represent the node and hyperedge representations as $\mathbf{P} \in \mathbb{R}^{|V|\times d}$ and $\mathbf{Q} \in \mathbb{R}^{|E|\times d}$, respectively, where each row $p_v$ ($q_e$) represents the $d$-dimensional embedding vector of node $v$ (hyperedge $e$).

\begin{table}[t]
\centering
\caption{Notations and their descriptions}
\setlength\tabcolsep{8pt}
\begin{tabular}{cl}
\toprule
 \textbf{Notation} & \textbf{Description}\\
\midrule
$H$ & a hypergraph that consists of nodes and hyperedges \\
$V, E$ & the set of nodes, the set of hyperedges \\
$X$ & the input node features \\
\midrule
$\mathbf{H}$ & the incidence matrix of $H$ \\
$\mathbf{D}^V, \mathbf{D}^E$ & the degree matrices of nodes and hyperedges \\
$\mathbf{P}, \mathbf{Q}$ & the node and hyperedge representations  \\
\midrule
$f(\cdot)$ & a hypergraph encoder \\
$agg(\cdot)$ & a node aggregator for hyperedge candidates \\
$pred(\cdot)$ & a hyperedge predictor \\
\midrule
$p_f, p_m$ & the node feature and membership masking rates \\
$\beta$ & the weight of the auxiliary task \\
$\mathcal{L}(\cdot)$ & loss function \\
$\mathbf{W}, b$ & the learnable weight and bias matrices \\

\bottomrule
\end{tabular}
\label{table:notations}
\end{table}

\vspace{1.5mm}
\subsubsection{\textbf{Problem definition}}
This work aims to solve the \textit{hyperedge prediction} problem, which is formally defined as follows.

\vspace{1mm}
\noindent
\textbf{\textsc{Problem 1} (\textsc{Hyperedge prediction}).}
Given a hypergraph $\mathbf{H}\in \{0,1\}^{|V|\times |E|}$, node feature $\mathbf{X}\in \mathbb{R}^{|V|\times F}$, and a hyperedge candidate $e' \notin E$,
the goal of hyperedge prediction is to predict whether $e'$ is real or not.

\vspace{1mm}

The process of hyperedge prediction is two-fold~\cite{hwang2022ahp}:

\noindent
(1) \textbf{Hypergraph encoding}: 
a hypergraph encoder, $f: (\mathbb{R}^{|V|\times F}, \mathbb{R}^{|V|\times |E|})$ $\rightarrow (\mathbb{R}^{|V|\times d}, \mathbb{R}^{|E|\times d})$, produces the node and hyperedge embeddings based on the observed hypergraph structure, i.e., $f(\mathbf{X}, \mathbf{H})=(\mathbf{P},\mathbf{Q})$.

\noindent
(2) \textbf{Hyperedge candidate scoring}: a node aggregator, $agg: \mathbb{R}^{|e'| \times d} \rightarrow \mathbb{R}^{d}$, produces the single embedding of a given hyperedge candidate $e'$ by aggregating the embeddings of nodes in $e'$;
finally, the aggregated embedding is fed into a predictor, $pred: \mathbb{R}^{d} \rightarrow \mathbb{R}^{1}$, to compute the probability of the hyperedge candidate $e'$ being formed.

\begin{figure*}[t]
    \centering
    \includegraphics[width=0.95\linewidth]{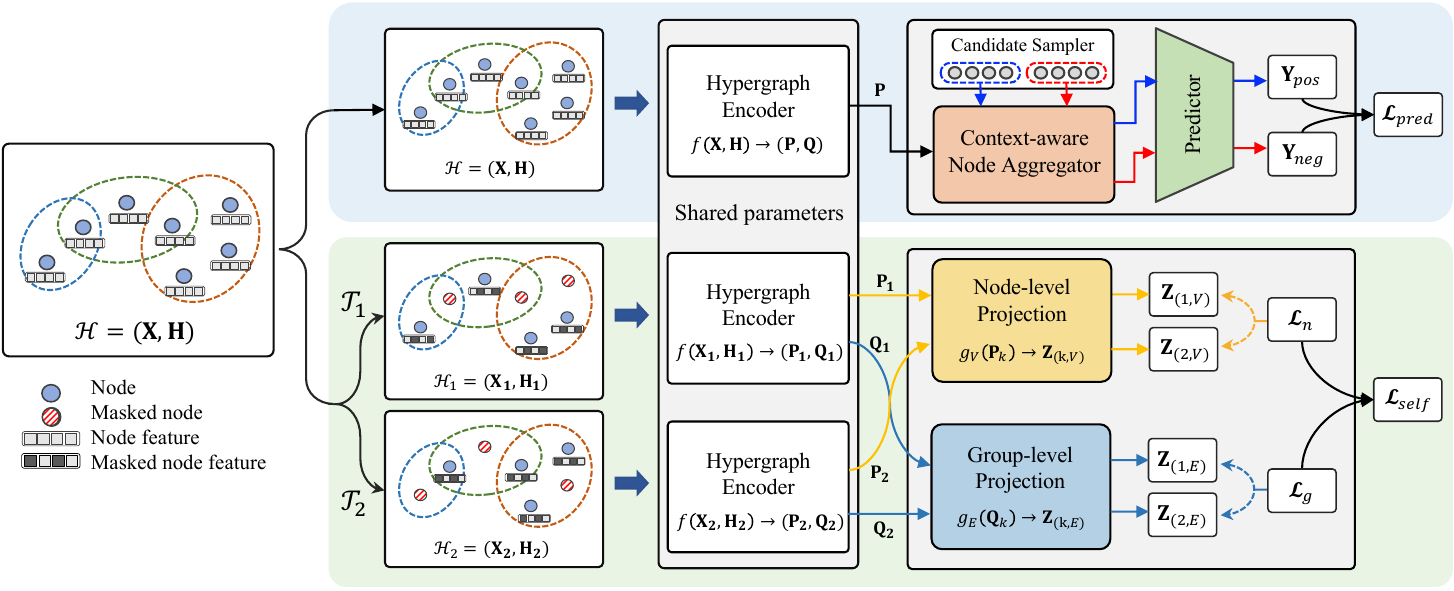}
    \vspace{-3mm}
    \caption{The overview of {\m}: (1) Context-aware hyperedge prediction (\textcolor{bleudefrance}{upper}) and (2) Self-supervised contrative hypergraph learning (\textcolor{mygreen}{lower}).}
    \label{fig:proposed-overview}
    \vspace{-3mm}
\end{figure*}

Based on this process, we aim to train the hypergraph encoder $f(\cdot)$, the node aggregator $agg(\cdot)$, and the hyperedge predictor $pred(\cdot)$ 
that minimize the loss $\mathcal{L}$ in an \textit{end-to-end} way.
Note that we define the loss function $\mathcal{L}(\cdot)$ based on two tasks, i.e., \textit{hyperedge prediction} as a primary task and \textit{self-supervised contrastive learning} as an auxiliary task as illustrated in Figure~\ref{fig:proposed-overview}.
We will describe the details of $f(\cdot)$, $agg(\cdot)$, $pred(\cdot)$, and $\mathcal{L}(\cdot)$ in Sections~\ref{sec:proposed-primiary} and~\ref{sec:proposed-auxiliary}.

% \vspace{-2mm}
\subsection{Context-Aware Hyperedge Prediction}\label{sec:proposed-primiary}

As illustrated in Figure~\ref{fig:proposed-overview}, {\m} jointly tackles two tasks: 
\textit{hyperedge prediction} as a primary task (\textcolor{bleudefrance}{upper}) and \textit{self-supervised contrastive learning} as an auxiliary task (\textcolor{mygreen}{lower}).
In this section, we explain how {\m} addresses the hyperedge prediction, which consists of (1) hypergraph encoding and (2) hyperedge candidate scoring.

For (1) hypergraph encoding, as {\m} is \textit{agnostic} to hypergraph encoders, any hypergraph neural networks (HGNNs)~\cite{dong2020hnhn,yadati2019hypergcn,feng2019hypergraph,chien2021you,wu2022hypergraph,ding2020more}, producing the node representations ($\mathbf{P}$) to be used for (2) hyperedge candidate scoring, could be applied to {\m}.
As we explained before, however, clique-expansion-based HGNNs~\cite{dong2020hnhn,yadati2019hypergcn} are unable to fully capture group-wise relations.
Thus, to better learn group-wise relations, 
we carefully design a hypergraph encoder of {\m} based on a \textit{2-stage aggregation strategy} (i.e., node-to-hyperedge and hyperedge-to-node aggregation) by following~\cite{dong2020hnhn,chien2021you,wu2022hypergraph} (\textbf{Section~\ref{sec:proposed-primiary-encoding}}).

For (2) hyperedge candidate scoring (our main focus), 
we point out a critical challenge that has been largely under-explored yet: 
\textbf{(C1) Node aggregation}. ``\textit{How to aggregate the nodes in each hyperedge candidate for accurate hyperedge prediction?}"
Naturally, in real-world scenarios, group-wise relations among objects are formed in a very complex manner.
In protein-protein interaction (PPI) networks, for example,
a group-wise relation among an arbitrary number of proteins could be formed only when the proteins co-induce a single chemical reaction together, 
where each protein may have a different degree of influence to its group-wise relation (i.e., the chemical reaction).
% While, a pair-wise relation is formed when two proteins co-induce the same reaction.
Thus, for predicting unobserved hyperedges (e.g., new chemical reaction) accurately,
\textit{it is crucial to precisely capture the degrees of influences in the complex and subtle relation among the nodes} that would form a hyperedge.

Existing hyperedge prediction methods~\cite{yadati2020nhp,hwang2022ahp}, however, simply aggregate a group of nodes without considering the complex relations (e.g, average pooling),
which degrades the accuracy of hyperedge prediction eventually.
From this motivation, we propose a method of \textit{context-aware node aggregation} that computes different degrees of influences of nodes in a hyperedge candidate to its formation and produces ``the \textit{context-aware} embedding'' of the hyperedge candidate, by aggregating the node embeddings based on their influences (\textbf{Section~\ref{sec:proposed-primiary-scoring}}).

\subsubsection{\textbf{Hypergraph encoding}}\label{sec:proposed-primiary-encoding}
In this step, a hypergraph encoder $f(\cdot)$ produces the node and hyperedge embeddings via a 2-stage aggregation strategy~\cite{dong2020hnhn,chien2021you,wu2022hypergraph}.
Specifically, {\m} updates each \textit{hyperedge embedding} by aggregating 
the embeddings of its incident nodes, 
$f_{V\rightarrow E}: \mathbb{R}^{|V|\times d} \rightarrow \mathbb{R}^{|E|\times d}$ (i.e., node-to-hyperedge aggregation),
and then updates each \textit{node embedding} by aggregating the embeddings of the hyperedges that it belongs to,
$f_{E\rightarrow V}: \mathbb{R}^{|E|\times d} \rightarrow \mathbb{R}^{|V|\times d}$ (i.e., hyperedge-to-node aggregation).
This 2-stage process is repeated by the number of layers $k$ of the hypergraph encoder model.
Formally, given a hypergraph incidence matrix $\mathbf{H}$ and an input node feature matrix $\mathbf{X}$,
the node and hyperedge embeddings at the \textit{k}-th layer, $\mathbf{P}^{(k)}$ and $\mathbf{Q}^{(k)}$, are defined as:
\begin{align}
    \mathbf{Q}^{(k)} &=  \sigma(\mathbf{D}^{-1}_{E}\mathbf{H}^{T}\mathbf{P}^{(k-1)} \mathbf{W}^{(k)}_{E} + b^{(k)}_{E}), \\ 
    \mathbf{P}^{(k)} &= \sigma(\mathbf{D}^{-1}_{V}\mathbf{H}\mathbf{Q}^{(k)} \mathbf{W}^{(k)}_{V} + b^{(k)}_{V}),
    \label{eq:hgnn}
\end{align}
where $\mathbf{P}^{(0)}=\mathbf{X}$, 
$\mathbf{W}^{(k)}_*$ and $b^{(k)}_*$ are trainable weight and bias matrices, respectively, 
$\mathbf{D}^{-1}_*$ is the normalization term,
and $\sigma$ is a non-linear activation function (PReLU~\cite{he2015delving}).
As illustrated in Figure~\ref{fig:proposed-overview}, the weights and biases of the hypergraph encoder ($\mathbf{W}_*$ and $b_*$) are shared in the self-supervised learning part.

It is worth noting that more-complicated neural network models~\cite{ding2020more,chien2021you,yang2022semi,wu2022hypergraph} could be adopted as the hypergraph encoder of our {\m} since our method is \textit{agnostic} to the hypergraph encoder architecture.

\subsubsection{\textbf{Hyperedge candidate scoring}}\label{sec:proposed-primiary-scoring}
In this step, given the learned node embeddings $\mathbf{P}$ and a hyperedge candidate $e'$,  
(1) a node aggregator, $agg: \mathbb{R}^{|e'|\times d} \rightarrow \mathbb{R}^d$, produces $q^*_{e'}$, the embedding of the hyperedge candidate $e'$,
and (2) a predictor $pred: \mathbb{R}^{d} \rightarrow \mathbb{R}^1$ computes the probability of the hyperedge candidate $e'$ being formed based on $q^*_{e'}$.

\vspace{1.5mm}
\noindent
\textbf{Context-aware hyperedge prediction.}
To reflect the different degrees of nodes' influences on a hyperedge candidate in its node aggregation,
we devise a simple but effective node aggregation method, i.e., \textit{context-aware} node aggregator, $agg(\cdot)$.
We first calculate the relative degrees of influences of the nodes in a hyperedge candidate to its formation by using the attention mechanism~\cite{vaswani2017attention},
and update each node embedding based on the relative degrees of influences.
Formally, given a hyperedge candidate $e'=\{v'_{1}, v'_{2}, ..., v'_{|{e'}|}\}$ and the learned embeddings of the nodes in hyperedge candidate $e'$, $\mathbf{P}[e',:]\in \mathbb{R}^{|e'|\times d}$,
the \textit{influence-reflected} embedding of node $v'_{j}$, $p^*_{v'_j}$, and the relative influence of $v'_i$ to $v'_j$, $\alpha_{i,j}$, are defined as:
\begin{align}
    p^*_{v'_j} &= \sum_{v'_i \in {e'}}{\alpha_{i,j}\cdot p_{v'_i} \mathbf{W}^{'}_{agg}}, \\
    \alpha_{i,j} &= \frac{exp(p_{v'_i}\mathbf{W}^{''}_{agg}\cdot x^{\top})}{\sum_{{v'_j}\in e'}exp(p_{v'_j}\mathbf{W}^{''}_{agg}\cdot x^{\top})},
    \label{eq:attention}
\end{align}
where $\mathbf{W}^{'}_{agg}, \mathbf{W}^{''}_{agg} \in \mathbb{R}^{d\times d}$ and $x \in \mathbb{R}^d$ are trainable parameters.

Then, we aggregate the influence-reflected embeddings of the nodes in hyperedge candidate $e'$, $\mathbf{P}^*[e',:]$,
via \textit{element-wise max pooling} to filter the important contextual information of each node,
finally computing the probability of $e'$ being formed, $\hat{y}_{e'}$, as:
\begin{align}
    \hat{y}_{e'} = pred(q^*_{e'}), \hspace{2mm} q^*_{e'} = MaxPool(\mathbf{P}^*[e',:]),
    \label{eq:maxpoolin}
\end{align}
where $q^*_{e'} \in \mathbb{R}^{d}$ is the final embedding of hyperedge candidate $e'$,
which can reflect the complex and subtle relation among the nodes of the hyperedge candidate,
and $pred(\cdot)$ is a hyperedge predictor (a fully-connected layer ($d \times 1$), followed by a sigmoid function).

To the best of our knowledge, this is the first work to adopt the attention-based method to aggregate the nodes in a hyperedge candidate for accurate hyperedge prediction.
We will empirically show the effectiveness of our context-aware node aggregation method in Section~\ref{sec:eval-result-eq2-ablation}.

\vspace{1.5mm}
\noindent
\textbf{Model training.}
For the model training and validation of {\m}, 
we consider both positive and negative examples (i.e., existing and non-existing hyperedges).
Specifically, to sample negative examples,
we use the following heuristic negative sampling (NS) methods~\cite{patil2020negative}, each of which has the different degrees of difficulty:
\begin{itemize}[leftmargin=10pt]
    \item \textbf{Sized NS (SNS)}: sampling $k$ random nodes (easy).
    \item \textbf{Motif NS (MNS)}: sampling a $k$-connected component in a clique-expanded hypergraph (difficult).
    \item \textbf{Clique NS (CNS)}: selecting a hyperedge $e$ and replacing one of its incident nodes $u\in e$
    with a node $v\notin e$, which is linked to all the other incident nodes, i.e., ($e \setminus \{u\}) \cup \{v\}$ (most difficult).
\end{itemize}

\noindent
Thus, we aim to train the model parameters of {\m} so that positive examples obtain higher scores while negative examples obtain lower scores.
Formally, given a set $E'$ of hyperedge candidates, the prediction loss is defined as:
\begin{align}
    \mathcal{L}_{pred} = -\frac{1}{|E'|} \sum_{e' \in E'}
    \underbrace{y_{e'}\cdot \log{\hat{y}_{e'}}}_{\text{\sf\footnotesize positives}} + \underbrace{(1-y_{e'})\cdot \log{(1-\hat{y}_{e'})}}_{\text{\sf\footnotesize negatives}},
    \label{eq:bceloss}
\end{align}
where $y_{e'}$ is the label of the hyperedge candidate $e'$ (1 or 0).

\subsection{Self-Supervised Hypergraph Learning}\label{sec:proposed-auxiliary}
In real-world hypergraphs,
there exist only a small number of group-wise relations~\cite{yu2021self,zhang2021double}.
This inherent data sparsity problem makes it very challenging to precisely capture group-wise relations among nodes, 
which often results in the accuracy degradation in hyperedge prediction.
To alleviate \textbf{(C2) the data sparsity} problem in hyperedge prediction, 
we incorporate the \textit{self-supervised contrastive learning}~\cite{you2020graph,zhu2020deep,lee2022m} in the training process of {\m} (See Figure~\ref{fig:proposed-overview}),
which provides complementary information to better learn group-wise relations among nodes in a hypergraph.

A general process of contrastive learning is as follows:
(1) generating two augmented views of a given hypergraph and (2) training the model parameters to minimize the contrast between the two views.
There are two important questions to answer in contrastive learning:
\textbf{(Q1)} ``\textit{How to generate two augmented views to fully exploit the latent semantics behind the original hypergraph?}" and
\textbf{(Q2)} ``\textit{What to contrast between the two augmented views?}"
To answer these questions,
we (1) propose a \textit{hyperedge-aware augmentation} method that generates two augmented views, preserving the structural properties of the original hypergraph for (Q1) (\textbf{Section~\ref{sec:proposed-auxiliary-augment}})
and (2) consider both node-level and group-level contrasts in constructing the training loss (i.e., \textit{dual contrastive loss}) for (Q2) (\textbf{Section~\ref{sec:proposed-auxiliary-contrast}}).

\subsubsection{\textbf{Hypergraph augmentation}}\label{sec:proposed-auxiliary-augment}
\begin{figure}[t]
    \centering
    \includegraphics[width=1.0\linewidth]{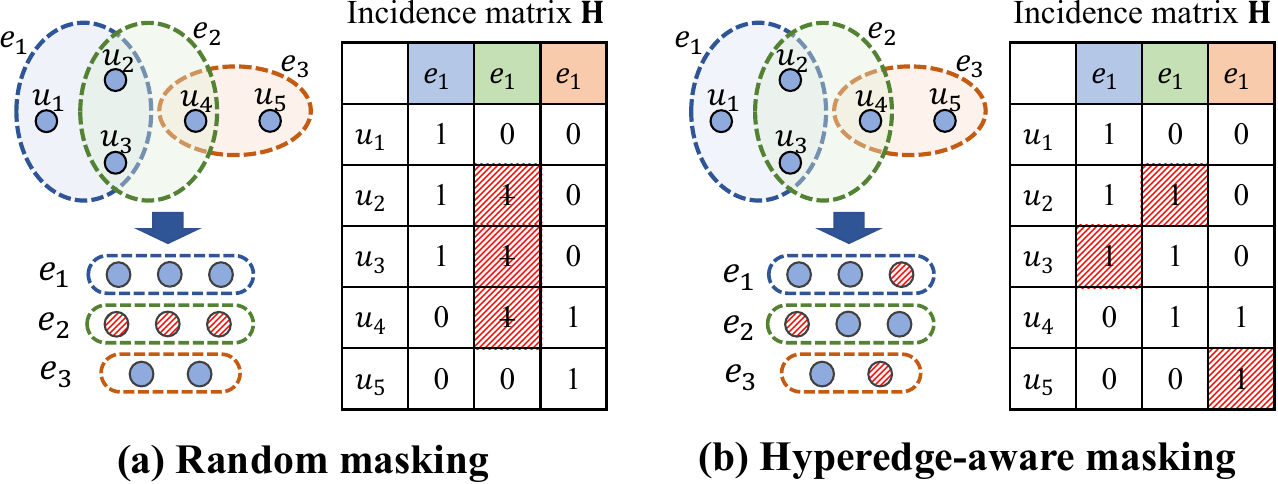}
    % \vspace{-3mm}
    \caption{Comparison of (a) random membership masking with (b) our hyperedge-aware membership masking.}
    % \vspace{-5mm}
    \label{fig:proposed-augment}
\end{figure}

In contrastive learning, generating augmented views is crucial since the latent semantics of the original hypergraph to capture could be different depending on the views. 
Despite its importance, the hypergraph augmentation still remains largely under-explored.
Existing works~\cite{you2020graph,zhu2020deep,lee2022m}, however, adopt a simple random augmentation method that generates augmented views by (i) directly dropping hyperedges or (ii) masking random nodes (members) in hyperedges (i.e., \textit{random membership masking}).
Specifically, it uses a random binary mask of the size $S=nnz(\mathbf{H})$, where $nnz(\mathbf{H})$ is the number of non-zero elements in a hypergraph incidence matrix $\mathbf{H}$.
It might happen that a majority (or all) of members might be masked in \textit{some} hyperedges, while only a few (or none of) members are masked in others.
Figure~\ref{fig:proposed-augment}(a) shows a toy example that \textit{all} members in hyperedge $e_2$ are masked (i.e., the group-wise relation disappears), 
while members in hyperedges $e_1$ and $e_3$ are not masked at all.
Therefore, this random augmentation method may impair the original hypergraph structure,
which results in decreasing the effect of contrastive learning eventually.

From this motivation,
we argue that \textit{it is critical to generate augmented views that preserve the original hypergraph structure} (e.g., the distribution of hyperedges).
To this end, we propose a simple yet effective augmentation method that generates two augmented views, considering variable sizes of hyperedges for (Q1).
Specifically, our method masks random $p_m\%$ members of each hyperedge \textit{individually} (i.e., \textit{hyperedge-aware membership masking}), 
rather than masking $p_m\%$ members of all hyperedges at once.
Thus, as shown in Figure~\ref{fig:proposed-augment}(b),
all existing group-wise relations of the original hypergraph can be preserved in the augmented views.
This implies that our hyperedge-aware method is able to successfully preserve the structural properties of the original hypergraph, 
which enables {\m} to fully exploit the latent semantics behind the original hypergrpah.

We also employ \textit{random node feature masking} by following~\cite{lee2022m,you2020graph,zhu2020deep}.
For node feature masking, we mask random $p_f\%$ dimensions of node features.
As a result, {\m} generates two augmented views of a hypergraph, $\mathcal{H}_1=(\mathbf{X}_1, \mathbf{H}_1)$ and $\mathcal{H}_2=(\mathbf{X}_2, \mathbf{H}_2)$.
Algorithm~\ref{algo:augment} shows the entire process of our hyperedge-aware augmentation.
We will evaluate our hyperedge-aware augmentation method and its hyperparameter sensitivity to $p_m$ and $p_f$ in Sections~\ref{sec:eval-result-eq2-ablation} and \ref{sec:eval-result-eq3-sensitivity}, respectively.

\begin{algorithm}[t]
\small
\DontPrintSemicolon 
\SetVlineSkip{1pt}
\SetInd{0.1em}{1.5em}
\SetKwInOut{Input}{\hspace{0.7em}Input}
\SetKwInOut{Output}{Output}
\Input{Node features $\mathbf{X}$, hypergraph $\mathbf{H}=(V,E)$, membership and feature masking rates $p_m$ and $p_f$}
\Output{Augmented hypergraph $\mathcal{H}^*$}

\SetKwFunction{FMain}{HyperedgeAwareAugment}
\SetKwProg{Pn}{Function}{:}{\KwRet}
\Pn{\FMain{$\mathbf{X}$, $\mathbf{H}$, $p_m$, $p_f$}}{
    $V^*\leftarrow \emptyset$, $E^*\leftarrow \emptyset$, $d \leftarrow |\mathbf{X}[0]|$, $\text{\sf FeatureMask}\leftarrow []$  \\
    \For(\tcp*[f]{1. Membership masking}){$e_j \in E$}{
        $e^* \leftarrow \emptyset$ \tcp*[f]{Masked hyperedge} \\
        \For{$v_i \in e$} {
            \If{$S\sim \mathcal{B}(1-p_m)$}{
                $e^* \leftarrow e^*\cup \{v_i\}$, $V^* \leftarrow V^*\cup \{v_i\}$
            }
        }
        $E^* \leftarrow E^*\cup \{e^*\}$
    }
    $\mathbf{X}^* \leftarrow \mathbf{X}[V^*,:]$, $\mathbf{H}^* \leftarrow (V^*, E^*)$ \\
    \For(\tcp*[f]{2. Node feature masking}){$i=1 \rightarrow d$}{
        \If(\tcp*[f]{Generating FeatureMask}){$S\sim \mathcal{B}(1-p_f)$}{
            $\text{\sf FeatureMask}.\text{\sf append}(1)$ 
        }
        \Else{
            $\text{\sf FeatureMask}.\text{\sf append}(0)$ 
        }
    }
    $\mathbf{X}^* \leftarrow \mathbf{X}^* \otimes \text{\sf FeatureMask}$ \tcp*[f]{Applying FeatureMask}\\
    \Return{$\mathcal{H}^* \leftarrow (\mathbf{X}^*, \mathbf{H}^*)$}    
}
\textbf{end function}
\caption{{\sc Hyperedge-Aware Augmentation}}\label{algo:augment}
\end{algorithm}

\subsubsection{\textbf{Hypergraph contrastive learning}}\label{sec:proposed-auxiliary-contrast}
For the two augmented views, $\mathcal{H}_1=(\mathbf{X}_1,\mathbf{H}_1)$ and $\mathcal{H}_2=(\mathbf{X}_2,\mathbf{H}_2)$,
we produce the node and hyperedge embeddings, $\mathbf{P}_{i}$ and $\mathbf{Q}_{i}$, respectively, where $i=1,2$ for each augmented view.
We use the same hypergraph encoder $f(\cdot)$ as explained in Section~\ref{sec:proposed-primiary-encoding}.
Then, 
we apply node and hyperedge projectors, $g_{V}: \mathbb{R}^{|V|\times d} \rightarrow \mathbb{R}^{|V|\times d}$ and $g_{E}: \mathbb{R}^{|E|\times d} \rightarrow \mathbb{R}^{|E|\times d}$,
to the learned node and hyperedge embeddings ($\mathbf{P}_{i}$ and $\mathbf{Q}_{i}$), in order to represent them to better fit the form in constructing the contrastive loss by following~\cite{chen2020simple}.
Thus, given the learned node and hyperedge embeddings for the $i$-th augmented view, $\mathbf{P}_{i}$ and $\mathbf{Q}_{i}$,
their projected embeddings, $\mathbf{Z}_{(i,V)}$ and $\mathbf{Z}_{(i,E)}$, are defined as:
\begin{align}
    \mathbf{Z}_{(i,V)} = g_{V}(\mathbf{P}_{i}), \hspace{1em}
    \mathbf{Z}_{(i,E)} = g_{E}(\mathbf{Q}_{i}).
    \label{eq:projection}
\end{align}
As the projector $g_*(\cdot)$, we use a two-layer MLP model ($d\times d_{proj}\times d$) with the ELU non-linear function~\cite{clevert2015fast}.

Then, based on the projected node and hyperedge embeddings, $\mathbf{Z}_{(i,V)}$ and $\mathbf{Z}_{(i,E)}$, 
we measure the contrast between the two contrastive views.
We consider not only the \textit{node-level} but also \textit{group-level} contrasts as self-supervisory signals in constructing the contrastive loss, i.e., \textit{dual contrastive loss}, for (Q2).
These dual contrastive signals are complementary information to better learn both node-level and group-level structural information of the original hypergraph,
thereby improving the accuracy in hyperedge prediction (i.e., alleviating (C2) the data sparsity problem).

Formally, 
given the projected node and hyperedge embeddings for each augmented view, $\mathbf{Z}_{(i,V)}$ and $\mathbf{Z}_{(i,E)}$,
the contrastive loss with dual contrasts is defined as (See the yellow and blue dotted lines in Figure~\ref{fig:proposed-overview}):
\begin{align}
    \mathcal{L}_{con} = &-\underbrace{\log{sim(\mathbf{Z}_{(1,V)}, \mathbf{Z}_{(2,V)})}}_{\text{\sf\footnotesize node-level contrast}} \nonumber \\
    &- \underbrace{\log{sim(\mathbf{Z}_{(1,E)}, \mathbf{Z}_{(2,E)})}}_{\text{\sf\footnotesize group-level contrast}},
    \label{eq:loss-contrastive}
\end{align}
where $sim(\cdot)$ is the cosine similarity used as a similarity function in {\m}.
Finally, we unify the two losses of the hyperedge prediction (primary task) and self-supervised contrastive learning (auxiliary task) by a weighted sum. 
Thus, the unified loss of {\m} is finally defined as:
\begin{align}
    \mathcal{L} = \mathcal{L}_{pred} + \beta\mathcal{L}_{con},
    \label{eq:loss-total}
\end{align}
where $\beta$ is a hyperparameter to control the weight of the auxiliary task.
Accordingly, all model parameters of {\m} are trained to jointly optimize the two tasks.
We will evaluate the impact of the hyperparameter $\beta$ on the accuracy of {\m} in Section~\ref{sec:eval-result-eq3-sensitivity}.

% summary
As a result, 
{\m} effectively addresses the two important but under-explored challenges of hyperedge prediction by employing two strategies:
(1) \textit{context-aware node aggregation} that considers the complex relation among nodes that would form a hyperedge for (C1) and
(2) \textit{self-supervised contrastive learning} that provides complementary information to better learn group-wise relations for (C2).

% \vspace{-2mm}
\subsection{Complexity Analysis}\label{sec:proposed-complexity}
% We analyze the space and time complexity of {\m}.

\vspace{1.5mm}
\noindent
\textbf{Space complexity.}
{\m} consists of (1) a hypergraph encoder, (2) a node aggregator, (3) a hyperedge predictor, and (4) a projector. 
The parameter size of a hypergraph encoder is $d\times d\times k\times 2$, 
where $d$ is the embedding dimensionality and $k$ is the number of layers. % (we use only one layer in this work).
The parameter sizes of a node aggregator, a hyperedge predictor, and a projector are $d\times d\times 3$, $d$, and $d\times d\times 2$, respectively.
In addition,
the space for node and hyperedge embeddings, $|V|\times d$ and $|E|\times d$, is commonly required in any hyperedge prediction methods. 
Thus, since $k$ is much smaller than $d$, $|V|$, and $|E|$, 
the overall space complexity of {\m} is $O((|V|+|E|+d)\cdot d)$, i.e., linear to the hypergraph size.
As a result, the space complexity of {\m} is comparable to those of existing methods 
since the additional space for our context-aware node aggregator and projector is much smaller than the commonly required space (i.e., $|V|+|E| \gg d$).

\vspace{1.5mm}
\noindent
\textbf{Time complexity.}
% \subsubsection{\textbf{Tsime complexity}}
The computational overhead of {\m} comes from (1) hypergraph encoding, (2) node aggregation, (3) hyperedge prediction, (4) projection, and (5) contrast.
The computational overhead of hypergraph encoding is $O(d\times |\mathbf{H}|\times k\times 2)$, 
where $|\mathbf{H}|$ is the number of non-zero elements in the hypergraph incidence matrix.
The context-aware node aggregation requires the time complexity of $O(d^2\times |e'|\times 3)$, 
where $|e'|$ is the size of a hyperedge candidate $e'$.
The overheads of hyperedge prediction and projection are $O(d)$ and $O(d^2\times 2)$, respectively.
Finally, the contrast overhead is $O(|V|+|E|)\cdot d$ (i.e., node-level and group-level).
Thus, the overall time complexity of {\m} is $O(|\mathbf{H}|+d+|V|+|E|)\cdot d$, i.e., linear to the hypergraph size,
since $|e'|$ and $k$ are much smaller than $d$, $|V|$, $|E|$, and $|\mathbf{H}|$, 
where we note the first term (the common overhead of hypergraph encoding $O(|\mathbf{H}|\cdot d)$) is dominant.
This implies that the time complexity of {\m} is also comparable to those of existing hyperedge prediction methods.
We will evaluate the scalability of {\m} with the increasing size of hypergraphs in Section~\ref{sec:eval-result-eq4-scalability}.
\section{Experimental Validation}\label{sec:eval}

In this section, we comprehensively evaluate {\m} by answering the following evaluation questions (EQs): 
\begin{itemize}[leftmargin=8pt]
    \item \textbf{EQ1 (Accuracy)}. To what extent does {\m} improve the existing hyperedge prediction methods in terms of the accuracy in hyperedge prediction?
    \item \textbf{EQ2 (Ablation study)}. How does each of our proposed strategies contributes to the model accuracy of {\m}?
    \item \textbf{EQ3 (Sensitivity)}. How sensitive is the model accuracy of {\m} to the hyperparameters ($\beta$, $p_f$ and $p_m$)?
    \item \textbf{EQ4 (Efficiency)}. How efficient is the model training of {\m}, compared to the existing methods?
    \item \textbf{EQ5 (Scalability)}. How does the training of {\m} scale up with the increasing size of hypergraphs?
\end{itemize}

\subsection{Experimental Setup}\label{sec:eval-setup}
\noindent
\textbf{Datasets.} 
We use six real-world hypergraphs (Table~\ref{table:datasets}), which were also used in~\cite{yadati2019hypergcn,dong2020hnhn,hwang2022ahp}:
(1) three co-citation datasets (\textsf{Citeseer}, \textsf{Cora}, and \textsf{Pubmed}), 
(2) two authorship datasets (\textsf{Cora-A} and \textsf{DBLP-A}), 
and (3) one collaboration dataset (\textsf{DBLP}).
In the co-citation datasets, each node indicates a paper and each hyperedge indicates the set of papers co-cited by a paper;
in the authorship dataset, each node indicates a paper and each hyperedge indicates the set of papers written by an author;
in the collaboration dataset, each node indicates a researcher and each hyperedge indicates the set of researchers who wrote the same paper.
For all the datasets,
we use the bag-of-word features from the abstract of each paper as in~\cite{hwang2022ahp}.
% Table~\ref{table:datasets} shows the data statistics.

\begin{table}[ht]
\small
\centering
\caption{Statistics of hypergraph datasets}
% \vspace{-3mm}
\label{table:datasets}
\setlength\tabcolsep{7pt}
\def\arraystretch{1.0} % row space
\begin{tabular}{c||r|r|r|cc}
\toprule
 
Dataset & $|V|$ & $|E|$ & \# Features & Type \\
% \multirow{2}{*}{Dataset} & \multirow{2}{*}{Nodes} & \multirow{2}{*}{Hyperedges} & \multirow{2}{*}{Features} & Size of hyperedges\\
%  & & & & Avg. & Max. \\

\midrule
\textbf{\textsf{Citeseer}} & 1,457 & 1,078 & 3,703 & Co-citation \\
\textbf{\textsf{Cora}}     & 1,434 & 1,579 & 1,433 & Co-citation \\
\textbf{\textsf{Pubmed}}   & 3,840 & 7,962 & 500 & Co-citation \\
\midrule
\textbf{\textsf{Cora-A}}    & 2,388 & 1,072 & 1,433 & Authorship \\
\textbf{\textsf{DBLP-A}}    & 39,283 & 16,483 & 4,543 & Authorship \\
\midrule
\textbf{\textsf{DBLP}}   & 15,639 & 22,964 & 4,543 & Collaboration \\

\bottomrule
\end{tabular}
\end{table}

\begin{table*}[t]
\centering

\caption{Hyperedge prediction accuracy on six real-world hypergraphs. {\m} consistently outperforms all competitors in terms of both AUROC and AP averaged over four test sets (The bold font indicates the best result on each test set).}\label{table:eval-accuracy}
\setlength\tabcolsep{5pt} % column space
\def\arraystretch{0.95} % row space
\begin{tabular}{cc||cccc|c|cccc|c}
\toprule

\multirow{2}{*}{Dataset} & Metric & \multicolumn{5}{c}{AUROC} & \multicolumn{5}{c}{Average Precision (AP)} \\ 
\cmidrule(lr){3-7} \cmidrule(lr){8-12} 
& Test set & SNS & MNS & CNS & MIX & Average & SNS & MNS & CNS & MIX & Average \\

\midrule
\midrule
\multirow{6}{*}{\rotatebox{90}{\textbf{{Citeseer}}}}
& Expansion     & 0.663 & 0.781 & 0.331 & 0.588 & 0.591 $\pm$ 0.011 & 0.765 & 0.817 & 0.498 & 0.630 & 0.681 $\pm$ 0.001 \\
& HyperSAGNN    & 0.540 & 0.410 & 0.473 & 0.478 & 0.475 $\pm$ 0.019 & 0.627 & 0.455 & 0.497 & 0.507 & 0.512 $\pm$ 0.015  \\
& NHP           & \textbf{\underline{0.991}} & 0.701 & 0.510 & 0.817 & 0.751 $\pm$ 0.009 & \textbf{\underline{0.990}} & 0.731 & 0.520 & 0.768 & 0.751 $\pm$ 0.011  \\
& AHP           & 0.943 & \underline{0.881} & \underline{0.651} & \underline{0.820} & \underline{0.824 $\pm$ 0.020} & 0.952 & \underline{0.870} & \underline{0.660} & \underline{0.795} & \underline{0.819 $\pm$ 0.022}  \\
& \textbf{{\m}} & 0.925 & \textbf{0.921} & \textbf{0.720} & \textbf{0.857} & \textbf{0.856 $\pm$ 0.011} & 0.928 & \textbf{0.919} & \textbf{0.701} & \textbf{0.831} & \textbf{0.845 $\pm$ 0.009}  \\
\cmidrule(lr){2-12}
& Improvement (\%) & \textcolor{red}{-6.65\%} & \textcolor{blue}{+4.54\%} & \textcolor{blue}{+10.60\%} & \textcolor{blue}{+4.51\%} & \textcolor{blue}{+3.88\%} & \textcolor{red}{-6.26\%} & \textcolor{blue}{5.63\%} & \textcolor{blue}{+6.21\%} & \textcolor{blue}{+4.53\%} & \textcolor{blue}{+3.17\%} \\

\midrule
\multirow{6}{*}{\rotatebox{90}{\textbf{{Cora}}}}
& Expansion     & 0.470 & 0.707 & 0.256 & 0.476 & 0.477 $\pm$ 0.009 & 0.637 & 0.764 & 0.454 & 0.563 & 0.607 $\pm$ 0.009 \\
& HyperSAGNN    & 0.617 & 0.527 & 0.494 & 0.540 & 0.545 $\pm$ 0.021 & 0.687 & 0.574 & 0.508 & 0.566 & 0.584 $\pm$ 0.019  \\
& NHP           & 0.943 & 0.641 & 0.472 & 0.774 & 0.703 $\pm$ 0.015 & 0.949 & 0.678 & 0.509 & \underline{0.744} & 0.718 $\pm$ 0.020  \\
& AHP           & \textbf{\underline{0.964}} & \underline{0.860} & \underline{0.572} & \underline{0.799} & \underline{0.799 $\pm$ 0.019} & \textbf{\underline{0.961}} & \underline{0.837} & \underline{0.552} & 0.740 & \underline{0.772 $\pm$ 0.035}  \\
& \textbf{{\m}} & 0.923 & \textbf{0.867} & \textbf{0.671} & \textbf{0.824} & \textbf{0.822 $\pm$ 0.011} & 0.915 & \textbf{0.854} & \textbf{0.644} & \textbf{0.789} & \textbf{0.801 $\pm$ 0.016}  \\
\cmidrule(lr){2-12}
& Improvement (\%) & \textcolor{red}{-4.25\%} & \textcolor{blue}{+0.81\%} & \textcolor{blue}{+17.31\%} & \textcolor{blue}{+3.13\%} & \textcolor{blue}{+2.88\%} & \textcolor{red}{-4.79\%} & \textcolor{blue}{+2.03\%} & \textcolor{blue}{+16.67\%} & \textcolor{blue}{+6.62\%} & \textcolor{blue}{+3.76\%} \\

\midrule
\multirow{6}{*}{\rotatebox{90}{\textbf{{Pubmed}}}}
& Expansion     & 0.520 & 0.730 & 0.241 & 0.497 & 0.497 $\pm$ 0.015 & 0.675 & 0.755 & 0.440 & 0.565 & 0.612 $\pm$ 0.010 \\
& HyperSAGNN    & 0.525 & 0.686 & 0.546 & 0.580 & 0.584 $\pm$ 0.066 & 0.534 & 0.680 & \underline{0.529} & 0.561 & 0.576 $\pm$ 0.050 \\
& NHP           & \textbf{\underline{0.973}} & 0.694 & 0.524 & 0.745 & 0.733 $\pm$ 0.004 & \textbf{\underline{0.973}} & 0.656 & 0.513 & 0.678 & 0.707 $\pm$ 0.004 \\
& AHP           & 0.917 & \underline{0.840} & \underline{0.553} & \underline{0.763} & \underline{0.763 $\pm$ 0.009} & 0.918 & \underline{0.834} & 0.526 & \underline{0.717} & \underline{0.749 $\pm$ 0.007} \\
& \textbf{{\m}} & 0.805 & \textbf{0.871} & \textbf{0.640} & \textbf{0.772} & \textbf{0.772 $\pm$ 0.009} & 0.810 & \textbf{0.880} & \textbf{0.644} & \textbf{0.765} & \textbf{0.775 $\pm$ 0.008} \\
\cmidrule(lr){2-12}
& Improvement (\%) & \textcolor{red}{-17.26\%} & \textcolor{blue}{+3.69\%} & \textcolor{blue}{15.73\%} & \textcolor{blue}{+1.18\%} & \textcolor{blue}{+1.18\%} & \textcolor{red}{-16.75\%} & \textcolor{blue}{+5.52\%} & \textcolor{blue}{+21.74\%} & \textcolor{blue}{+6.69\%} & \textcolor{blue}{+3.47\%} \\

\midrule
\multirow{6}{*}{\rotatebox{90}{\textbf{{Cora-A}}}}
& Expansion     & 0.690 & 0.842 & 0.434 & 0.658 & 0.656 $\pm$ 0.011 & 0.690 & 0.876 & 0.577 & 0.672 & 0.706 $\pm$ 0.020 \\
& HyperSAGNN    & 0.386 & 0.591 & 0.542 & 0.505 & 0.506 $\pm$ 0.019 & 0.532 & 0.643 & 0.545 & 0.563 & 0.571 $\pm$ 0.009 \\
& NHP           & 0.909 & 0.672 & 0.550 & 0.773 & 0.723 $\pm$ 0.015 & 0.925 & 0.720 & 0.585 & 0.766 & 0.748 $\pm$ 0.019 \\
& AHP           & \underline{0.958} & \underline{0.924} & \underline{0.782} & \underline{0.887} & \underline{0.888 $\pm$ 0.014} & \underline{0.957} & \underline{0.898} & \underline{0.796} & \underline{0.878} & \underline{0.882 $\pm$ 0.014} \\
& \textbf{{\m}} & \textbf{0.971} & \textbf{0.975} & \textbf{0.833} & \textbf{0.931} & \textbf{0.927 $\pm$ 0.011} & \textbf{0.969} & \textbf{0.973} & \textbf{0.832} & \textbf{0.926} & \textbf{0.925 $\pm$ 0.011} \\
\cmidrule(lr){2-12}
& Improvement (\%) & \textcolor{blue}{+1.36\%} & \textcolor{blue}{+5.52\%} & \textcolor{blue}{+6.52\%} & \textcolor{blue}{+4.96\%} & \textcolor{blue}{+4.39\%} & \textcolor{blue}{+1.25\%} & \textcolor{blue}{+8.35\%} & \textcolor{blue}{+4.52\%} & \textcolor{blue}{+5.47\%} & \textcolor{blue}{+4.88\%} \\

\midrule
\multirow{6}{*}{\rotatebox{90}{\textbf{{DBLP-A}}}}
& Expansion     & 0.634 & 0.826 & 0.350 & 0.603 & 0.603 $\pm$ 0.006 & 0.730 & 0.852 & 0.512 & 0.641 & 0.687 $\pm$ 0.004 \\
& HyperSAGNN    & 0.548 & 0.791 & 0.563 & 0.636 & 0.634 $\pm$ 0.007 & 0.686 & 0.805 & 0.552 & 0.655 & 0.675 $\pm$ 0.004 \\
& NHP           & \textbf{\underline{0.966}} & 0.623 & 0.555 & 0.721 & 0.716 $\pm$ 0.005 & \textbf{\underline{0.965}} & 0.604 & 0.534 & 0.663 & 0.693 $\pm$ 0.007 \\
& AHP           & 0.916 & \underline{0.926} & \underline{0.668} & \underline{0.838} & \underline{0.837 $\pm$ 0.004} & 0.928 & \underline{0.928} & \underline{0.707} & \underline{0.836} & \underline{0.850 $\pm$ 0.003} \\
& \textbf{{\m}} & 0.929 & \textbf{0.957} & \textbf{0.747} & \textbf{0.877} & \textbf{0.877 $\pm$ 0.003} & 0.933 & \textbf{0.955} & \textbf{0.741} & \textbf{0.863} & \textbf{0.873 $\pm$ 0.005} \\
\cmidrule(lr){2-12}
& Improvement (\%) & \textcolor{red}{-3.83\%} & \textcolor{blue}{+3.35\%} & \textcolor{blue}{+11.83\%} & \textcolor{blue}{+4.65\%} & \textcolor{blue}{+4.78\%} & \textcolor{red}{-3.32\%} & \textcolor{blue}{+2.91\%} & \textcolor{blue}{+4.81\%} & \textcolor{blue}{+3.23\%} & \textcolor{blue}{+2.71\%} \\

\midrule
\multirow{6}{*}{\rotatebox{90}{\textbf{{DBLP}}}}
& Expansion     & 0.645 & 0.801 & 0.366 & 0.607 & 0.607 $\pm$ 0.005 & 0.751 & \underline{\textbf{0.856}} & 0.518 & 0.655 & 0.698 $\pm$ 0.004 \\
& HyperSAGNN    & 0.448 & 0.574 & \underline{0.572} & 0.530 & 0.531 $\pm$ 0.018 & 0.562 & 0.602 & \underline{0.586} & 0.577 & 0.582 $\pm$ 0.016 \\
& NHP           & 0.663 & 0.540 & 0.503 & 0.572 & 0.569 $\pm$ 0.003 & 0.608 & 0.523 & 0.501 & 0.542 & 0.544 $\pm$ 0.002 \\
& AHP           & \underline{\textbf{0.946}} & \underline{0.820} & 0.568 & \underline{0.778} & \underline{0.778 $\pm$ 0.002} & \underline{\textbf{0.947}} & 0.815 & 0.561 & \underline{0.735} & \underline{0.764 $\pm$ 0.007} \\
& \textbf{{\m}} & 0.875 & \textbf{0.836} & \textbf{0.708} & \textbf{0.807} & \textbf{0.807 $\pm$ 0.015} & 0.874 & 0.832 & \textbf{0.696} & \textbf{0.793} & \textbf{0.799 $\pm$ 0.011} \\
\cmidrule(lr){2-12}
& Improvement (\%) & \textcolor{red}{-7.50\%} & \textcolor{blue}{+1.95\%} & \textcolor{blue}{+23.78\%} & \textcolor{blue}{+3.73\%} & \textcolor{blue}{+3.73\%} & \textcolor{red}{-7.70\%} & \textcolor{red}{-2.80\%} & \textcolor{blue}{+18.77\%} & \textcolor{blue}{+7.89\%} & \textcolor{blue}{+4.58\%} \\

\bottomrule
\end{tabular}
\end{table*}

\vspace{1mm}
\noindent
\textbf{Evaluation protocol.}
We evaluate {\m} by using the protocol exactly same as that used in~\cite{hwang2022ahp}.
For each dataset, we use five data splits,
where hyperedges (i.e., positive examples) in each split are randomly divided into the training (60\%), validation (20\%), and test (20\%) sets.
To comprehensively evaluate {\m}, we use four different validation and test sets, 
each of which has different negative examples with various degrees of difficulty, as in~\cite{hwang2022ahp}.
Specifically, we (1) sample negative examples as many as positive examples by using four heuristic negative sampling (NS) methods~\cite{patil2020negative}, which are explained in Section~\ref{sec:proposed-primiary} (i.e., sized NS (SNS), motif NS (MNS), clique NS (CNS), and a mixed one (MIX)), and (2) add them to each validation/test set (i.e., the ratio of positives to negatives is 1:1).
As evaluation metrics, we use AUROC (area under the ROC curve) and AP (average precision), 
where higher values of these metrics indicate higher hyperedge prediction accuracy.
Then, we (1) measure AUROC and AP on each test set at the epoch when the averaged AUROC over the four validation sets is maximized, 
and (2) report the averaged AUROC and AP on each test set over five runs.
All datasets and their splits used in this paper are available at: \myhref{\codelink}{\codelink}.

\vspace{1mm}
\noindent
\textbf{Competing methods.}
We compare {\m} with the following four hyperedge prediction methods in our experiments.
\begin{itemize}[leftmargin=10pt]
    \item \textbf{Expansion}~\cite{yoon2020expansion}: Expansion represents a hypergraph via multiple \textit{n}-projected graphs and predicts future hyperedges based on the multiple projected graphs.

    \item \textbf{HyperSAGNN}~\cite{zhang2020hypersagnn}: HyperSAGNN employs a self-attention based GNN model to learn hyperedges with variable sizes and predicts whether each hyperedge candidate is formed.
    
    \item \textbf{NHP}~\cite{yadati2020nhp}: NHP applies hyperedge-aware GCNs to a hypergraph to learn the node embeddings and aggregates the embeddings of nodes in each hyperedge candidate by using max-min pooling. 

    \item \textbf{AHP}~\cite{hwang2022ahp}: AHP, a state-of-the-art method, employs an adversarial-training-based model to generate negative examples for use in the model training and employs max-min pooling for the aggregation of the nodes in a hyperedge candidate.
    
\end{itemize}
For all competing methods, we use their results reported in~\cite{hwang2022ahp} since we follow the exactly same evaluation protocol and use the exactly same data splits as in~\cite{hwang2022ahp}.

\vspace{1mm}
\noindent
\textbf{Implementation details.} 
We implement {\m} by using PyTorch 1.11 and Deep Graph Library (DGL) 0.9 on Ubuntu 20.04.
We run all experiments on the machine equipped with an Intel i7-9700k CPU with 64GB main memory and two NVIDIA RTX 2080 Ti GPUs, each of which has 11GB memory and is installed with CUDA 11.3 and cuDNN 8.2.1.
For all datasets, we set the batch size as 32 to fully utilize the GPU memory and the dimensionality of node and hyperedge embeddings as 512, following~\cite{hwang2022ahp,dong2020hnhn}.
For the model training, we use the Adam optimizer~\cite{kingma2015adam} with the learning rate $\eta=\text{5e-3}$ and the weight decay factor 5e-4 for all datasets.
We use the motif NS (MNS)\footnote{We have also tried to use other negative samplers in the training of {\m} but have observed that their impacts on the accuracy are negligible.}~\cite{patil2020negative} to select negative hyperedges in the model training with the ratio of positive examples to negative examples as 1:1 (i.e., 32 positives and 32 negatives are used in each iteration).
For self-supervised learning, 
we adjust the control factor of the auxiliary task, $\beta$, from 0.0 to 1.0, 
and the node feature masking and membership masking rates, $p_f$ and $p_m$, from 0.1 to 0.9 in step of 0.1 for all datasets, 
which will be elaborated in Section~\ref{sec:eval-result-eq3-sensitivity}.

\begin{table*}[t]
\centering
\caption{Effects of the proposed strategies in improving the accuracy of {\m}. Each of our strategies is always beneficial to improving the accuracy of {\m} in hyperedge prediction (The bold font indicates the best result on each test set).}\label{table:ablation}

\setlength\tabcolsep{4.4pt} % column space
\def\arraystretch{0.95} % row space
\begin{tabular}{cc||cccc|c|cccc|c}
\toprule

\multirow{2}{*}{Dataset} & Metric & \multicolumn{5}{c}{AUROC} & \multicolumn{5}{c}{Average Precision (AP)} \\ 
\cmidrule(lr){3-7} \cmidrule(lr){8-12} 
& Test set & SNS & MNS & CNS & MIX & Average & SNS & MNS & CNS & MIX & Average \\

\midrule
\midrule
\multirow{6}{*}{\rotatebox{90}{\textbf{{Citeseer}}}}
& \textbf{{\m}}-\textcolor{red}{No} & 0.878 & 0.847 & 0.630 & 0.786 & 0.786 $\pm$ 0.003 & 0.890 & 0.841 & 0.653 & 0.775 & 0.790 $\pm$ 0.007  \\
& \textbf{{\m}}-\textcolor{brown}{CL} & 0.907 & 0.890 & 0.679 & 0.832 & 0.827 $\pm$ 0.019 & 0.905 & 0.874 & 0.675 & 0.815 & 0.817 $\pm$ 0.013  \\
& \textbf{{\m}}-\textcolor{mygreen}{HCL} & 0.908 & 0.897 & 0.691 & 0.839 & 0.833 $\pm$ 0.013 & 0.909 & 0.880 & 0.691 & 0.824 & 0.826 $\pm$ 0.005  \\

& \textbf{{\m}}-\textcolor{blue}{ALL} & \textbf{0.925} & \textbf{0.921} & \textbf{0.720} & \textbf{0.857} & \textbf{0.856 $\pm$ 0.011} & \textbf{0.928} & \textbf{0.919} & \textbf{0.701} & \textbf{0.831} & \textbf{0.845 $\pm$ 0.009}  \\
\cmidrule(lr){2-12}
& Improvement (\%) & \textcolor{blue}{+5.35\%} & \textcolor{blue}{+8.74\%} & \textcolor{blue}{+14.29\%} & \textcolor{blue}{+9.03\%} & \textcolor{blue}{+8.91\%} & \textcolor{blue}{+4.27\%} & \textcolor{blue}{+9.27\%} & \textcolor{blue}{+7.35\%} & \textcolor{blue}{+7.23\%} & \textcolor{blue}{+6.96\%} \\

\midrule
\multirow{6}{*}{\rotatebox{90}{\textbf{{Cora}}}}
& \textbf{{\m}}-\textcolor{red}{No} & 0.852 & 0.750 & 0.532 & 0.711 & 0.712 $\pm$ 0.019 & 0.856 & 0.759 & 0.531 & 0.684 & 0.707 $\pm$ 0.021  \\
& \textbf{{\m}}-\textcolor{brown}{CL} & 0.895 & 0.837 & 0.600 & 0.782 & 0.779 $\pm$ 0.015 & 0.873 & 0.809 & 0.566 & 0.727 & 0.744 $\pm$ 0.019  \\
& \textbf{{\m}}-\textcolor{mygreen}{HCL} & 0.893 & 0.835 & 0.600 & 0.780 & 0.777 $\pm$ 0.017 & 0.879 & 0.816 & 0.565 & 0.730 & 0.747 $\pm$ 0.018  \\

& \textbf{{\m}}-\textcolor{blue}{ALL} & \textbf{0.923} & \textbf{0.867} & \textbf{0.671} & \textbf{0.824} & \textbf{0.822 $\pm$ 0.011} & \textbf{0.915} & \textbf{0.854} & \textbf{0.644} & \textbf{0.789} & \textbf{0.801 $\pm$ 0.016}  \\
\cmidrule(lr){2-12}
& Improvement (\%) & \textcolor{blue}{+8.33\%} & \textcolor{blue}{+15.60\%} & \textcolor{blue}{+26.13\%} & \textcolor{blue}{+15.89\%} & \textcolor{blue}{+15.45\%} & \textcolor{blue}{+6.89\%} & \textcolor{blue}{+12.52\%} & \textcolor{blue}{+21.28\%} & \textcolor{blue}{+15.35\%} & \textcolor{blue}{+13.30\%} \\

\midrule
\multirow{6}{*}{\rotatebox{90}{\textbf{{Pubmed}}}}
& \textbf{{\m}}-\textcolor{red}{No} & 0.782 & 0.844 & 0.558 & 0.727 & 0.728 $\pm$ 0.007 & 0.802 & 0.852 & 0.555 & 0.708 & 0.730 $\pm$ 0.007  \\
& \textbf{{\m}}-\textcolor{brown}{CL} & 0.806 & 0.845 & 0.562 & 0.735 & 0.737 $\pm$ 0.010 & 0.817 & 0.847 & 0.552 & 0.708 & 0.731 $\pm$ 0.007  \\
& \textbf{{\m}}-\textcolor{mygreen}{HCL} & \textbf{0.814} & 0.848 & 0.562 & 0.739 & 0.741 $\pm$ 0.008 & \textbf{0.823} & 0.851 & 0.547 & 0.708 & 0.732 $\pm$ 0.006  \\

& \textbf{{\m}}-\textcolor{blue}{ALL} & 0.805 & \textbf{0.871} & \textbf{0.640} & \textbf{0.772} & \textbf{0.772 $\pm$ 0.009} & 0.810 & \textbf{0.880} & \textbf{0.644} & \textbf{0.765} & \textbf{0.775 $\pm$ 0.008} \\
\cmidrule(lr){2-12}
& Improvement (\%) & \textcolor{blue}{+2.94\%} & \textcolor{blue}{+3.20\%} & \textcolor{blue}{+14.70\%} & \textcolor{blue}{+6.19\%} & \textcolor{blue}{+6.04\%} & \textcolor{blue}{+1.00\%} & \textcolor{blue}{+3.29\%} & \textcolor{blue}{+16.04\%} & \textcolor{blue}{+8.05\%} & \textcolor{blue}{+6.16\%} \\

\midrule
\multirow{6}{*}{\rotatebox{90}{\textbf{{Cora-A}}}}
& \textbf{{\m}}-\textcolor{red}{No} & 0.949 & 0.894 & 0.701 & 0.852 & 0.849 $\pm$ 0.020 & 0.951 & 0.906 & 0.738 & 0.857 & 0.863 $\pm$ 0.017  \\
& \textbf{{\m}}-\textcolor{brown}{CL} & 0.943 & 0.934 & 0.756 & 0.883 & 0.879 $\pm$ 0.030 & 0.944 & 0.936 & 0.772 & 0.881 & 0.884 $\pm$ 0.026  \\
& \textbf{{\m}}-\textcolor{mygreen}{HCL} & \textbf{0.972} & 0.949 & 0.833 & 0.921 & 0.919 $\pm$ 0.008 & \textbf{0.972} & 0.919 & \textbf{0.845} & 0.908 & 0.911 $\pm$ 0.007  \\

& \textbf{{\m}}-\textcolor{blue}{ALL} & 0.971 & \textbf{0.975} & \textbf{0.833} & \textbf{0.931} & \textbf{0.927 $\pm$ 0.011} & 0.969 & \textbf{0.973} & 0.832 & \textbf{0.926} & \textbf{0.925 $\pm$ 0.011} \\
\cmidrule(lr){2-12}
& Improvement (\%) & \textcolor{blue}{+2.32\%} & \textcolor{blue}{+9.06\%} & \textcolor{blue}{+18.83\%} & \textcolor{blue}{+9.27\%} & \textcolor{blue}{+9.19\%} & \textcolor{blue}{+1.89\%} & \textcolor{blue}{+7.40\%} & \textcolor{blue}{+12.74\%} & \textcolor{blue}{+8.05\%} & \textcolor{blue}{+7.18\%} \\

\bottomrule
\end{tabular}
\end{table*}

% \vspace{-3mm}

\subsection{Experimental Results}\label{sec:eval-result}
\subsubsection{\textbf{Accuracy (EQ1)}}\label{sec:eval-result-eq1-accuracy}
We first evaluate the hyperedge prediction accuracy of {\m}.
Table~\ref{table:eval-accuracy} shows the accuracies of all comparing methods in six real-world hypergraphs.
The results show that {\m} consistently outperforms \textit{all} competiting methods in \textit{all} datasets in both (averaged) AUROC and AP.
Specifically, {\m} achieves higher AUROC by up to 45.4\%, 38.3\%, 22.5\%, and 4.78\% than Expansion, HyperSAGNN, NHP, and AHP, respectively in \textsf{DBLP-A}.
We note that these improvements of {\m} over AHP (the best competitor) are remarkable, given that AHP has already improved other existing methods significantly in those datasets.
Consequently,
these results demonstrate that {\m} is able to effectively capture the group-wise relations among nodes by addressing the two challenges of hyperedge prediction successfully, i.e., (C1) \textit{node aggregation} and (C2) \textit{data sparsity},
through the proposed strategies:
(1) the context-aware node aggregation for (C1) and (2) the self-supervised learning with \textit{hyperedge-aware augmentation} and \textit{dual contrastive loss} for (C2).

Although {\m} is generally outperformed by the two best competitors (i.e., NHP and AHP) in the SNS setting, 
{\m} still achieves very high accuracies: 92.5\%, 92.3\%, 97.1\% (the best) and 92.9\% in the SNS setting of Citeseer, Cora, Cora-A, and DBLP-A, respectively. 
Examining the results of these two competitors more closely reveals that they show very low accuracies on the CNS test set (i.e., the most difficult test set), 
which is similar to or even worse than the accuracy of the random prediction ($\approx0.5$), 
while they achieve very high accuracies (almost perfect) in the SNS test set (i.e., the easiest test set). 
These accuracy gaps between the CNS and SNS test sets imply that they may be overfitting to the easy negative examples, thus which limits their ability to be generalized to other datasets. 
In other words, they do not successfully address the two challenges of hyperedge prediction that we identified – i.e., (C1) node aggregation and (C2) data sparsity – and thus fail to precisely capture the high-order information encoded in hyperedges.

On the other hand, {\m} consistently achieves high accuracies across all test settings including the SNS, MNS, CNS, and MIX settings.
Consequently, the accuracy differences among test settings are the smallest among all competing methods, which demonstrates that {\m} has superior generalization ability compared to all the competing methods.

\begin{figure*}[t]
\centering
\includegraphics[width=0.4\linewidth]{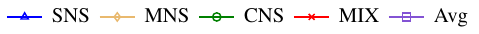}
\begin{tabular}{cccc}
    
    \includegraphics[width=0.22\linewidth]{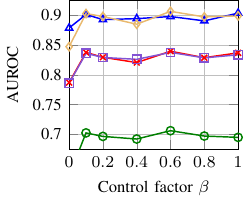}
     &  
    \includegraphics[width=0.22\linewidth]{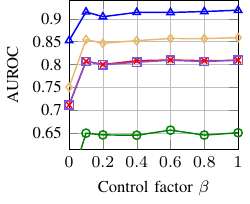}
    &
     \includegraphics[width=0.216\linewidth]{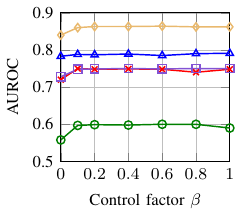}
     &
    \includegraphics[width=0.216\linewidth]{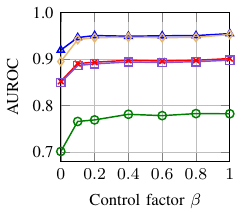}
     \\
    (a) {Citeseer}    
     & 
    (b) {Cora}
     &
    (c) {Pubmed}    
     & 
    (d) {Cora-A}
\end{tabular}

\caption{The impact of the dual contrative learning on the hyperedge prediction accuracy of {\m} according to the control hyperparameter $\beta$.
The auxiliary task is consistently beneficial to hyperedge prediction across a wide range of $\beta$ values.}\label{fig:eq3-contrastive}
\end{figure*}

\subsubsection{\textbf{Ablation study (EQ2)}}\label{sec:eval-result-eq2-ablation}
In this experiment, we verify the effectiveness of the proposed strategies of {\m} individually.
We compare the following four versions of {\m}:
\begin{itemize}[leftmargin=10pt]
    \item \textbf{{\m}}-\textcolor{red}{No}: the baseline version, excluding both strategies (i.e., neither context-aware node aggregation nor self-supervised contrastive learning). That is, in this version, node embeddings are generated using hypergraph neural networks, and the embeddings of nodes in a hyperedge candidate are aggregated via max-min pooling.    
    \item \textbf{{\m}}-\textcolor{brown}{CL}: the version with self-supervised contrastive learning with dual contrastive loss, but without hyperedge-aware augmentation and context-aware node aggregation.
    \item \textbf{{\m}}-\textcolor{mygreen}{HCL}: the version with self-supervised contrastive learning with dual contrastive loss and hyperedge-aware augmentation, but without context-aware node aggregation.
    \item \textbf{{\m}}-\textcolor{blue}{ALL}: the original version with \textit{all} strategies (i.e., context-aware node aggregation and self-supervised contrastive learning with dual contrastive loss and hyperedge-aware augmentation).
\end{itemize}
Table~\ref{table:ablation} shows the results of our ablation study.
Overall, each of our proposed strategies is always beneficial to improving the model accuracy of {\m}.
Specifically, when all strategies are applied to {\m} (i.e., {\m}-\textcolor{blue}{ALL}),
the averaged AUROC is improved by 8.91\%, 15.45\%, 6.04\%, and 9.19\% compared to the baseline (i.e., {\m}-\textcolor{red}{No}), in \textsf{Citeseer}, \textsf{Cora}, \textsf{Pubmed}, and \textsf{Cora}-A, respectively.
These results demonstrate that the two challenges of hyperedge prediction that we point out, i.e., (C1) node aggregation and (2) data sparsity, are critical for accurate hyperedge prediction and our proposed strategies employed in {\m} address them successfully.

Looking more closely,
\textbf{(1) effect of contrastive learning}: {\m}-\textcolor{brown}{CL} outperforms {\m}-\textcolor{red}{No} on all test sets of all datasets.
This result verifies the effect of the \textit{self-supervised contrastive learning} of {\m},
which alleviates (C2) the \textit{data sparsity} problem successfully by providing complementary information to better learn node and hyperedge representations, as we claimed in Section~\ref{sec:proposed-auxiliary}.
Then, \textbf{(2) effect of the hyperedge-aware augmentation}: {\m}-\textcolor{mygreen}{HCL} also improves {\m}-\textcolor{brown}{CL} consistently.
This demonstrates that our \textit{hyperedge-aware augmentation} method is more beneficial to hyperedge prediction than a simple random augmentation method. 
Thus, our method is able to generate two augmented views preserving the structural properties of the original hypergraph,
thereby enabling {\m} to fully exploit the latent semantics behind the original hypergraphs as we claimed in Section~\ref{sec:proposed-auxiliary-augment}.
Lastly, \textbf{(3) effect of the context-aware node aggregation}: {\m}-\textcolor{blue}{ALL} achieves higher accuracies than {\m}-\textcolor{mygreen}{HCL} in all datasets,
which verifies that our \textit{context-aware node aggregation} method is able to address (C1) the challenge of node aggregation effectively,
by capturing the complex and subtle relations among the nodes in a hyperedge candidate for accurate hyperedge prediction.

Note that the strategies of {\m} -- context-aware node aggregation and self-supervised learning with dual contrastive loss and hyperedge-aware augmentation -- can potentially be applied to other state-of-the-art methods. 
However, the second strategy (i.e., self-supervised learning with dual contrastive loss and hyperedge-aware augmentation) cannot be applied to other hyperedge prediction methods used in our experiments (AHP, Expansion, NHP, and HyperSAGNN) as it is specifically designed for hyperedge prediction methods that adopt a contrastive learning approach. 
It is worth noting that, to the best of our knowledge, {\m} is the first work to adopt contrastive learning for the hyperedge prediction problem.

\begin{figure}[t]
    \centering
    \begin{tabular}{c}
        \includegraphics[width=0.705\linewidth]{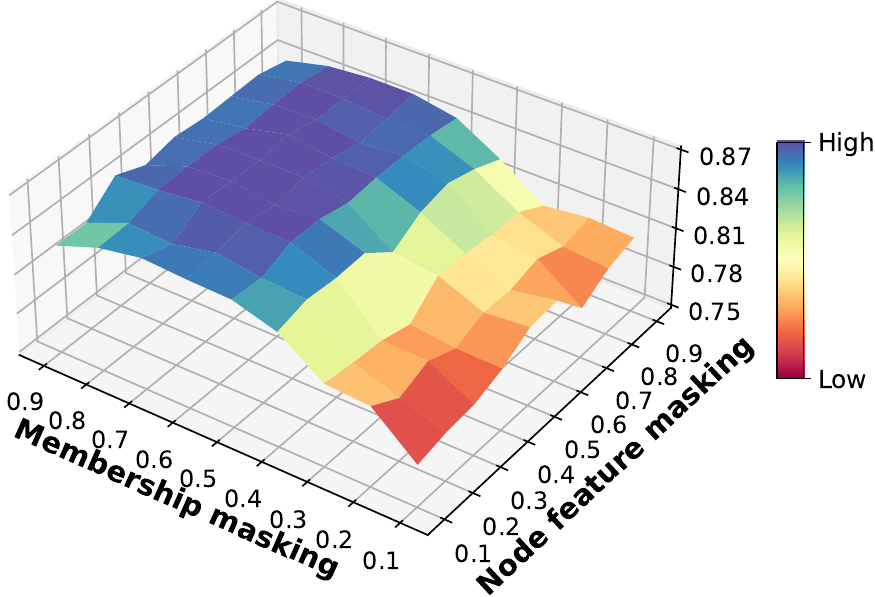}
        \\
        (a) Averaged AUROC on Citeseer
        \\
        \includegraphics[width=0.705\linewidth]{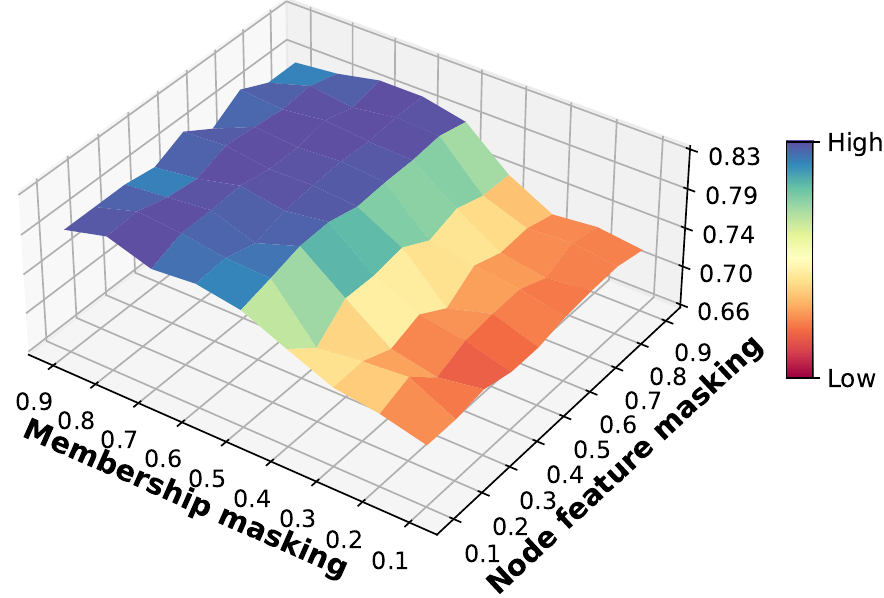}
        \\ 
         (b) Averaged AUROC on Cora
    \end{tabular}
    \caption{The hyperparameter sensitivity of {\m} to the membership and node feature masking rates $p_{m}$ and $p_{f}$. {\m} achieves high accuracy with a wide range of $p_{m}$ and $p_{f}$ values (i.e., the blue wide area on the surface).}\label{fig:eq3-masking}
\end{figure}

\subsubsection{\textbf{Sensitivity analysis (EQ3)}}\label{sec:eval-result-eq3-sensitivity}
In this experiment, we analyze the hyperparameter sensitivity of {\m}.
First, we evaluate the impact of the auxiliary task (i.e., self-supervised contrastive learning) on the model accuracy of {\m} according to the control factor $\beta$.
We measure the model accuracy of {\m} on four different test sets with varying $\beta$ from 0.0 (i.e., not used) to 1.0 (i.e., as the same as the primary task) in step of 0.1.
Figure~\ref{fig:eq3-contrastive} shows the results, where the \textit{x}-axis represents the control factor $\beta$ and the \textit{y}-axis represents the AUROC.
The model accuracy of {\m} is significantly improved in \textit{all} cases when $\beta$ is larger than 0.1, 
and {\m} achieves high prediction accuracy across a wide range of $\beta$ values ($\beta>=0.1$).
This result verifies that (i) self-supervised contrastive learning is consistently beneficial to improving the accuracy of {\m} by providing complementary information to better learn high-order information encoded in hyperedges 
and (ii) the accuracy of {\m} is insensitive to its hyperparameter $\beta$.

Then, we evaluate the impacts of the augmentation hyperparameters $p_m$ and $p_f$ on the model accuracy of {\m}.
As explained in Section~\ref{sec:proposed-auxiliary-augment}, 
the hyperparameter $p_m$ ($p_f$) controls how many members (dimensions) of each hyperedge (node feature vector) are masked in augmented views.
Thus, as $p_m$ ($p_f$) becomes larger, the more members (dimensions) of each hyperedge (node feature vector) are masked in augmented views.
We measure the model accuracy of {\m} with varying $p_m$ and $p_f$ from 0.1 to 0.9 in step of 0.1.
Figure~\ref{fig:eq3-masking} shows the results, where the \textit{x}-axis represents the membership masking rate $p_m$, the \textit{y}-axis represents the node feature masking rate $p_f$, and the \textit{z}-axis represents the averaged AUROC.
{\m} with $p_m$ above 0.4 consistently achieves higher accuracy than {\m} with $p_m$ below 0.4 regardless of $p_f$ (i.e., the blue wide area on the surface in Figure~\ref{fig:eq3-masking}).
On the other hand, {\m} with $p_m$ below 0.4 shows low hyperedge prediction accuracy (i.e., the red/orange area on the surface).
Specifically, {\m} with $p_m=0.1$ and $p_f=0.1$ (i.e., memberships and features are rarely masked) shows the worst result in the \textsf{Citeseer} dataset.
These results imply that
(1) \textit{the hyperedge membership masking is more important than the node feature masking in contrastive learning} that aims to capture the structural information of the original hypergraph 
and (2) {\m} is able to achieve high accuracy across a wide range of values of hyperparameters.
Based on these results, we believe that the accuracy of {\m} is \textit{insensitive} to the augmentation hyperparameters $p_m$ and $p_f$, and we recommend setting $p_m$ and $p_f$ as above $0.4$.

\begin{figure*}[t]
    \centering
    \includegraphics[width=0.9\linewidth]{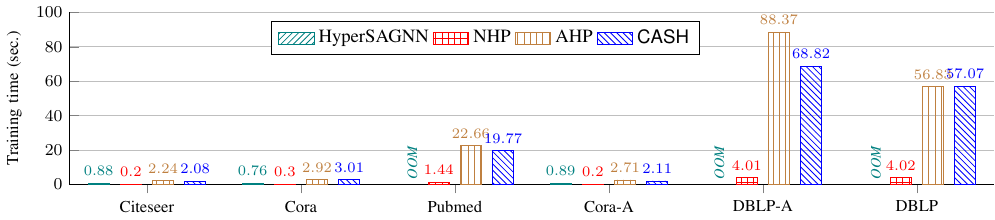}
    \caption{The average training time per epoch of hyperedge prediction methods on six real-world hypergraphs.}
    \vspace{-4mm}
    \label{fig:eq4-efficiency}
\end{figure*}

\subsubsection{\textbf{Efficiency (EQ4)}}\label{sec:eval-result-eq4-efficiency}
In this experiment, we compare {\m} with other methods in terms of training efficiency using six real-world hypergraphs. 
Specifically, we train {\m} and three competing methods (AHP, NHP, and HyperSAGNN) for 100 epochs, 
and measure the running time at every epoch. 
For the three competing methods, we use the official source codes provided by the authors and set the same values for their hyperparameters in this experiment.

Figure~\ref{fig:eq4-efficiency} shows the average running time per epoch of each method. 
First, {\m} completes the model training in a comparable time to or shorter time than AHP, the best competitor. 
Given that CASH consistently outperforms AHP in terms of the hyperedge prediction accuracy across all datasets and metrics (See Table~\ref{table:eval-accuracy}), this result implies that {\m} is able to capture high-order relations among real-world objects more precisely than AHP, with similar model complexity. 
We believe that this improvement of CASH over AHP is due to its effective handling of the two key challenges: (C1) node aggregation and (C2) data sparsity, through our proposed strategies of (1) context-aware node aggregation and (2) self-supervised contrastive learning.

Although NHP and HyperSAGNN complete their training in much shorter time than {\m}, {\m} significantly outperforms them in terms of the hyperedge prediction accuracy, achieving up to 38.3\% and 22.5\% higher AUROC than NHP and HyperSAGNN, respectively. 
Moreover, regarding the space complexity, HyperSAGNN fails to train on the Pubmed, DBLP-A, and DBLP datasets due to the out-of-memory (\textit{OOM}) issue. 
While, {\m} successfully trains on these large hypergraphs

\begin{figure*}[t]
\centering
\begin{tabular}{cccc}
    \centering
    \includegraphics[width=0.22\linewidth]{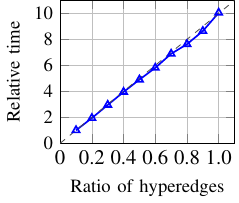}
     &  
    \includegraphics[width=0.22\linewidth]{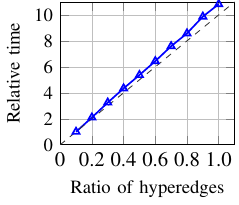}
     &
    \includegraphics[width=0.216\linewidth]{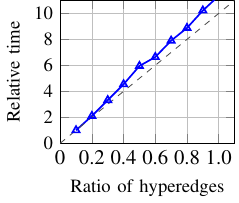}
     &
    \includegraphics[width=0.216\linewidth]{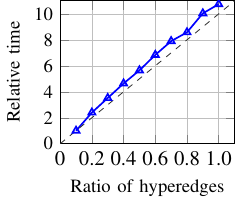}
     \\
    (a) {Citeseer}    
     & 
    (b) {Cora}
     &
    (c) {DBLP-A}    
     & 
    (d) {DBLP}
\end{tabular}

\caption{The training time per epoch of {\m} with the increasing number of hyperedges. {\m} provides (almost) linear scalability with the increasing number of hyperedges.}\label{fig:eq4-scalability}
\end{figure*}

\subsubsection{\textbf{Scalability (EQ5)}}\label{sec:eval-result-eq4-scalability}
Finally, we evaluate the scalability of {\m} in training with the increasing size of hypergraphs.
We train {\m} for 20 training epochs in four real-world hypergraphs -- two small hypergraphs (Citeseer and Cora) and two large hypergraphs (DBLP-A and DBLP), which are over 10 times larger than the smaller hypergraph -- varying the ratio of the training examples (i.e., hyperedges) from 10\% to 100\% in increments of 10\%, and measure the averaged training time per epoch. 
For brevity, we report the relative training time per epoch (i.e., a relative time of 1 represents the time per epoch for 10\% of the training examples in each dataset). 
Figure~\ref{fig:eq4-scalability} shows the results, where the \textit{x}-axis represents the ratio of training examples and the \textit{y}-axis represents the relative training time per epoch. 
The results reveal that the training of {\m} \textit{scales up linearly} with the increasing number of hyperedges,
which aligns with our theoretical analysis of the time complexity of {\m} as we explained in Section~\ref{sec:proposed-complexity}.

\section{Conclusion and Future Work}\label{sec:conclusion}
In this paper, 
we point out two important but under-explored challenges of hyperedge prediction, i.e., (C1) node aggregation and (C2) data sparsity.
To tackle the two challenges together, 
we propose a novel hyperedge prediction framework, named as {\m} that employs (1) the context-aware node aggregation for (C1) and (2) the self-supervised contrastive learning for (C2).
Furthermore, we propose the hyperedge-aware augmentation method to fully exploit the structural information of the original hypergraph and consider the dual contrasts to better capture the group-wise relations among nodes.
Via extensive experiments on six real-world hypergraphs,
we demonstrate that 
(1) (\textit{Accuracy}) {\m} consistently outperforms all competing methods in terms of the accuracy in hyperedge prediction, 
(2) (\textit{Effectiveness}) all proposed strategies are beneficial to improving the accuracy of {\m}, 
(3) (\textit{Insensitivity}) {\m} is able to achieve high accuracy across a wide range of values of hyperparameters (i.e., low hyperparameter sensitivity), 
and (4) (\textit{Efficiency}) {\m} completes the model training in a comparable time to or shorter time than the state-of-the-art method,
and (5) (\textit{Scalability}) {\m} provides almost linear scalability in training with the increasing size of hypergraphs.

\section*{Acknowledgments} 
This work was supported by Institute of Information \& Communications Technology Planning \& Evaluation (IITP) grant funded by the Korea government (MSIT) (RS-2022-00155586, 2022-0-00352, RS-2020-II201373).

% IITP
% RS-2022-00155586: SW 스타랩
% 2020-0-01373: 인공지능 대학원
% 2022-0-00352: OTT 추천


\bibliographystyle{IEEEtran}
% \balance
\bibliography{bibliography}

\begin{thebibliography}{10}
\providecommand{\url}[1]{#1}
\csname url@samestyle\endcsname
\providecommand{\newblock}{\relax}
\providecommand{\bibinfo}[2]{#2}
\providecommand{\BIBentrySTDinterwordspacing}{\spaceskip=0pt\relax}
\providecommand{\BIBentryALTinterwordstretchfactor}{4}
\providecommand{\BIBentryALTinterwordspacing}{\spaceskip=\fontdimen2\font plus
\BIBentryALTinterwordstretchfactor\fontdimen3\font minus
  \fontdimen4\font\relax}
\providecommand{\BIBforeignlanguage}[2]{{%
\expandafter\ifx\csname l@#1\endcsname\relax
\typeout{** WARNING: IEEEtran.bst: No hyphenation pattern has been}%
\typeout{** loaded for the language `#1'. Using the pattern for}%
\typeout{** the default language instead.}%
\else
\language=\csname l@#1\endcsname
\fi
#2}}
\providecommand{\BIBdecl}{\relax}
\BIBdecl

\bibitem{zhou2006learning}
D.~Zhou, J.~Huang, and B.~Sch{\"o}lkopf, ``Learning with hypergraphs:
  Clustering, classification, and embedding,'' \emph{the Advances in Neural
  Information Processing Systems (NeurIPS)}, 2006.

\bibitem{benson2018simplicial}
A.~R. Benson, R.~Abebe, M.~T. Schaub, A.~Jadbabaie, and J.~Kleinberg,
  ``Simplicial closure and higher-order link prediction,'' \emph{the National
  Academy of Sciences}, vol. 115, no.~48, pp. E11\,221--E11\,230, 2018.

\bibitem{do2020structural}
M.~T. Do, S.-e. Yoon, B.~Hooi, and K.~Shin, ``Structural patterns and
  generative models of real-world hypergraphs,'' in \emph{Proceedings of the
  ACM SIGKDD International Conference on Knowledge Discovery and Data Mining
  (KDD)}, 2020, pp. 176--186.

\bibitem{amburg2020clustering}
I.~Amburg, N.~Veldt, and A.~Benson, ``Clustering in graphs and hypergraphs with
  categorical edge labels,'' in \emph{Proceedings of The Web Conference (WWW)},
  2020, pp. 706--717.

\bibitem{comrie2021hypergraph}
C.~Comrie and J.~Kleinberg, ``Hypergraph ego-networks and their temporal
  evolution,'' in \emph{Proceedings of the IEEE International Conference on
  Data Mining (ICDM)}.\hskip 1em plus 0.5em minus 0.4em\relax IEEE, 2021, pp.
  91--100.

\bibitem{choe2022midas}
M.~Choe, J.~Yoo, G.~Lee, W.~Baek, U.~Kang, and K.~Shin, ``Midas: Representative
  sampling from real-world hypergraphs,'' in \emph{Proceedings of the ACM Web
  Conference (WWW)}, 2022, pp. 1080--1092.

\bibitem{jiang2018hyperx}
W.~Jiang, J.~Qi, J.~X. Yu, J.~Huang, and R.~Zhang, ``Hyperx: A scalable
  hypergraph framework,'' \emph{IEEE Transactions on Knowledge and Data
  Engineering (TKDE)}, vol.~31, no.~5, pp. 909--922, 2018.

\bibitem{jiang2022hypergraph}
P.~Jiang, X.~Deng, L.~Wang, Z.~Chen, and S.~Zhang, ``Hypergraph representation
  for detecting 3d objects from noisy point clouds,'' \emph{IEEE Transactions
  on Knowledge and Data Engineering (TKDE)}, 2022.

\bibitem{sun2023self}
X.~Sun, H.~Cheng, B.~Liu, J.~Li, H.~Chen, G.~Xu, and H.~Yin, ``Self-supervised
  hypergraph representation learning for sociological analysis,'' \emph{IEEE
  Transactions on Knowledge and Data Engineering (TKDE)}, 2023.

\bibitem{yoon2020expansion}
S.-e. Yoon, H.~Song, K.~Shin, and Y.~Yi, ``How much and when do we need
  higher-order information in hypergraphs? a case study on hyperedge
  prediction,'' in \emph{Proceedings of the ACM Web Conference (WWW)}, 2020,
  pp. 2627--2633.

\bibitem{hwang2022ahp}
H.~Hwang, S.~Lee, C.~Park, and K.~Shin, ``Ahp: Learning to negative sample for
  hyperedge prediction,'' in \emph{Proceedings of the International ACM SIGIR
  Conference on Research and Development in Information Retrieval (SIGIR)},
  2022, p. 2237–2242.

\bibitem{dong2020hnhn}
Y.~Dong, W.~Sawin, and Y.~Bengio, ``Hnhn: Hypergraph networks with hyperedge
  neurons,'' \emph{arXiv preprint arXiv:2006.12278}, 2020.

\bibitem{yadati2020nhp}
N.~Yadati, V.~Nitin, M.~Nimishakavi, P.~Yadav, A.~Louis, and P.~Talukdar,
  ``Nhp: Neural hypergraph link prediction,'' in \emph{Proceedings of the ACM
  International Conference on Information and Knowledge Management (CIKM)},
  2020, pp. 1705--1714.

\bibitem{feng2019hypergraph}
Y.~Feng, H.~You, Z.~Zhang, R.~Ji, and Y.~Gao, ``Hypergraph neural networks,''
  in \emph{Proceedings of the AAAI Conference on Artificial Intelligence
  (AAAI)}, vol.~33, no.~01, 2019, pp. 3558--3565.

\bibitem{chitra2019random}
U.~Chitra and B.~Raphael, ``Random walks on hypergraphs with edge-dependent
  vertex weights,'' in \emph{International conference on machine
  learning}.\hskip 1em plus 0.5em minus 0.4em\relax PMLR, 2019, pp. 1172--1181.

\bibitem{lee2022hashnwalk}
G.~Lee, M.~Choe, and K.~Shin, ``Hashnwalk: Hash and random walk based anomaly
  detection in hyperedge streams,'' \emph{arXiv preprint arXiv:2204.13822},
  2022.

\bibitem{xia2022hypergraph}
L.~Xia, C.~Huang, Y.~Xu, J.~Zhao, D.~Yin, and J.~Huang, ``Hypergraph
  contrastive collaborative filtering,'' in \emph{Proceedings of the
  International ACM SIGIR conference on Research and Development in Information
  Retrieval (SIGIR)}, 2022, pp. 70--79.

\bibitem{wang2020next}
J.~Wang, K.~Ding, L.~Hong, H.~Liu, and J.~Caverlee, ``Next-item recommendation
  with sequential hypergraphs,'' in \emph{Proceedings of the International ACM
  SIGIR Conference on Research and Development in Information Retrieval
  (SIGIR)}, 2020, pp. 1101--1110.

\bibitem{han2022dh}
J.~Han, Q.~Tao, Y.~Tang, and Y.~Xia, ``Dh-hgcn: Dual homogeneity hypergraph
  convolutional network for multiple social recommendations,'' in
  \emph{Proceedings of the International ACM SIGIR Conference on Research and
  Development in Information Retrieval (SIGIR)}, 2022, pp. 2190--2194.

\bibitem{li2022enhancing}
Y.~Li, C.~Gao, H.~Luo, D.~Jin, and Y.~Li, ``Enhancing hypergraph neural
  networks with intent disentanglement for session-based recommendation,'' in
  \emph{Proceedings of the International ACM SIGIR Conference on Research and
  Development in Information Retrieval (SIGIR)}, 2022, pp. 1997--2002.

\bibitem{yu2021self}
J.~Yu, H.~Yin, J.~Li, Q.~Wang, N.~Q.~V. Hung, and X.~Zhang, ``Self-supervised
  multi-channel hypergraph convolutional network for social recommendation,''
  in \emph{Proceedings of the Web conference (WWW)}, 2021, pp. 413--424.

\bibitem{zhang2021double}
J.~Zhang, M.~Gao, J.~Yu, L.~Guo, J.~Li, and H.~Yin, ``Double-scale
  self-supervised hypergraph learning for group recommendation,'' in
  \emph{Proceedings of the ACM International Conference on Information and
  Knowledge Management (CIKM)}, 2021, pp. 2557--2567.

\bibitem{xia2021self}
X.~Xia, H.~Yin, J.~Yu, Q.~Wang, L.~Cui, and X.~Zhang, ``Self-supervised
  hypergraph convolutional networks for session-based recommendation,'' in
  \emph{Proceedings of the AAAI Conference on Artificial Intelligence (AAAI)},
  vol.~35, no.~5, 2021, pp. 4503--4511.

\bibitem{liben2003link}
D.~Liben-Nowell and J.~Kleinberg, ``The link prediction problem for social
  networks,'' in \emph{Proceedings of the ACM International Conference on
  Information and Knowledge Management (CIKM)}, 2003, pp. 556--559.

\bibitem{lu2011link}
L.~L{\"u} and T.~Zhou, ``Link prediction in complex networks: A survey,''
  \emph{Physica A: Statistical Mechanics and its Applications}, vol. 390,
  no.~6, pp. 1150--1170, 2011.

\bibitem{li2013link}
D.~Li, Z.~Xu, S.~Li, and X.~Sun, ``Link prediction in social networks based on
  hypergraph,'' in \emph{Proceedings of the ACM Web Conference (WWW)}, 2013,
  pp. 41--42.

\bibitem{zhang2018beyond}
M.~Zhang, Z.~Cui, S.~Jiang, and Y.~Chen, ``Beyond link prediction: Predicting
  hyperlinks in adjacency space,'' in \emph{Proceedings of the AAAI Conference
  on Artificial Intelligence (AAAI)}, vol.~32, no.~1, 2018.

\bibitem{liu2017computational}
Y.~Liu, S.~Qiu, P.~Zhang, P.~Gong, F.~Wang, G.~Xue, and J.~Ye, ``Computational
  drug discovery with dyadic positive-unlabeled learning,'' in
  \emph{Proceedings of the SIAM International Conference on Data Mining
  (SDM)}.\hskip 1em plus 0.5em minus 0.4em\relax SIAM, 2017, pp. 45--53.

\bibitem{vaida2019hypergraph}
M.~Vaida and K.~Purcell, ``Hypergraph link prediction: Learning drug
  interaction networks embeddings,'' in \emph{Proceedings of the IEEE
  International Conference On Machine Learning And Applications (ICMLA)}.\hskip
  1em plus 0.5em minus 0.4em\relax IEEE, 2019, pp. 1860--1865.

\bibitem{tu2018structural}
K.~Tu, P.~Cui, X.~Wang, F.~Wang, and W.~Zhu, ``Structural deep embedding for
  hyper-networks,'' in \emph{Proceedings of the AAAI Conference on Artificial
  Intelligence (AAAI)}, vol.~32, no.~1, 2018.

\bibitem{yadati2019hypergcn}
N.~Yadati, M.~Nimishakavi, P.~Yadav, V.~Nitin, A.~Louis, and P.~Talukdar,
  ``Hypergcn: A new method for training graph convolutional networks on
  hypergraphs,'' \emph{Proceedings of the Advances in Neural Information
  Processing Systems (NeurIPS)}, 2019.

\bibitem{ding2020more}
K.~Ding, J.~Wang, J.~Li, D.~Li, and H.~Liu, ``Be more with less: Hypergraph
  attention networks for inductive text classification,'' in \emph{Proceedings
  of the Conference on Empirical Methods in Natural Language Processing
  (EMNLP)}.\hskip 1em plus 0.5em minus 0.4em\relax Association for
  Computational Linguistics (ACL), 2020, pp. 4927--4936.

\bibitem{yang2022semi}
C.~Yang, R.~Wang, S.~Yao, and T.~Abdelzaher, ``Semi-supervised hypergraph node
  classification on hypergraph line expansion,'' in \emph{Proceedings of the
  ACM International Conference on Information and Knowledge Management (CIKM)},
  2022, pp. 2352--2361.

\bibitem{wu2022hypergraph}
H.~Wu, Y.~Yan, and M.~K. Ng, ``Hypergraph collaborative network on vertices and
  hyperedges,'' \emph{IEEE Transactions on Pattern Analysis and Machine
  Intelligence (TPAMI)}, 2022.

\bibitem{chien2021you}
E.~Chien, C.~Pan, J.~Peng, and O.~Milenkovic, ``You are allset: A multiset
  function framework for hypergraph neural networks,'' \emph{arXiv preprint
  arXiv:2106.13264}, 2021.

\bibitem{zhang2020hypersagnn}
R.~Zhang, Y.~Zou, and J.~Ma, ``Hyper-sagnn: A self-attention based graph neural
  network for hypergraphs,'' in \emph{Proceedings of the International
  Conference on Learning Representations (ICLR)}, 2020.

\bibitem{nguyen2021centsmoothie}
D.~A. Nguyen, C.~H. Nguyen, and H.~Mamitsuka, ``Centsmoothie: Central-smoothing
  hypergraph neural networks for predicting drug-drug interactions,''
  \emph{arXiv preprint arXiv:2112.07837}, 2021.

\bibitem{patil2020negative}
P.~Patil, G.~Sharma, and M.~N. Murty, ``Negative sampling for hyperlink
  prediction in networks,'' in \emph{Proceedings of the Pacific-Asia Conference
  on Knowledge Discovery and Data Mining (PAKDD)}.\hskip 1em plus 0.5em minus
  0.4em\relax Springer, 2020, pp. 607--619.

\bibitem{ko2021mascot}
Y.~Ko, J.-S. Yu, H.-K. Bae, Y.~Park, D.~Lee, and S.-W. Kim, ``Mascot: A
  quantization framework for efficient matrix factorization in recommender
  systems,'' in \emph{2021 IEEE International Conference on Data Mining
  (ICDM)}.\hskip 1em plus 0.5em minus 0.4em\relax IEEE, 2021, pp. 290--299.

\bibitem{you2020graph}
Y.~You, T.~Chen, Y.~Sui, T.~Chen, Z.~Wang, and Y.~Shen, ``Graph contrastive
  learning with augmentations,'' \emph{Proceedings of the Advances in Neural
  Information Processing Systems (NeurIPS)}, vol.~33, pp. 5812--5823, 2020.

\bibitem{zhu2020deep}
Y.~Zhu, Y.~Xu, F.~Yu, Q.~Liu, S.~Wu, and L.~Wang, ``Deep graph contrastive
  representation learning,'' \emph{arXiv preprint arXiv:2006.04131}, 2020.

\bibitem{lee2022relational}
N.~Lee, D.~Hyun, J.~Lee, and C.~Park, ``Relational self-supervised learning on
  graphs,'' in \emph{Proceedings of the 31st ACM International Conference on
  Information and Knowledge Management (CIKM)}, 2022, pp. 1054--1063.

\bibitem{lee2022m}
D.~Lee and K.~Shin, ``I'm me, we're us, and i'm us: Tri-directional contrastive
  learning on hypergraphs,'' \emph{arXiv preprint arXiv:2206.04739}, 2022.

\bibitem{kipf2016semi}
T.~N. Kipf and M.~Welling, ``Semi-supervised classification with graph
  convolutional networks,'' \emph{arXiv preprint arXiv:1609.02907}, 2016.

\bibitem{du2021hypergraph}
B.~Du, C.~Yuan, R.~Barton, T.~Neiman, and H.~Tong, ``Hypergraph pre-training
  with graph neural networks,'' \emph{arXiv preprint arXiv:2105.10862}, 2021.

\bibitem{wei2022augmentations}
T.~Wei, Y.~You, T.~Chen, Y.~Shen, J.~He, and Z.~Wang, ``Augmentations in
  hypergraph contrastive learning: Fabricated and generative,'' \emph{Advances
  in Neural Information Processing Systems}, vol.~35, pp. 1909--1922, 2022.

\bibitem{alsentzer2020subgraph}
E.~Alsentzer, S.~Finlayson, M.~Li, and M.~Zitnik, ``Subgraph neural networks,''
  \emph{Advances in Neural Information Processing Systems (NeurIPS)}, vol.~33,
  pp. 8017--8029, 2020.

\bibitem{wang2021glass}
X.~Wang and M.~Zhang, ``Glass: Gnn with labeling tricks for subgraph
  representation learning,'' in \emph{International Conference on Learning
  Representations (ICLR)}, 2021.

\bibitem{hamidi2022subgraph}
R.~Hamidi~Rad, E.~Bagheri, M.~Kargar, D.~Srivastava, and J.~Szlichta,
  ``Subgraph representation learning for team mining,'' in \emph{Proceedings of
  the 14th ACM Web Science Conference 2022}, 2022, pp. 148--153.

\bibitem{he2015delving}
K.~He, X.~Zhang, S.~Ren, and J.~Sun, ``Delving deep into rectifiers: Surpassing
  human-level performance on imagenet classification,'' in \emph{Proceedings of
  the IEEE International Conference on Computer Vision (ICCV)}, 2015, pp.
  1026--1034.

\bibitem{vaswani2017attention}
A.~Vaswani, N.~Shazeer, N.~Parmar, J.~Uszkoreit, L.~Jones, A.~N. Gomez,
  {\L}.~Kaiser, and I.~Polosukhin, ``Attention is all you need,''
  \emph{Advances in Neural Information Processing Systems}, vol.~30, 2017.

\bibitem{chen2020simple}
T.~Chen, S.~Kornblith, M.~Norouzi, and G.~Hinton, ``A simple framework for
  contrastive learning of visual representations,'' in \emph{Proceedings of the
  IEEE International Conference On Machine Learning (ICML)}.\hskip 1em plus
  0.5em minus 0.4em\relax PMLR, 2020, pp. 1597--1607.

\bibitem{clevert2015fast}
D.-A. Clevert, T.~Unterthiner, and S.~Hochreiter, ``Fast and accurate deep
  network learning by exponential linear units (elus),'' \emph{arXiv preprint
  arXiv:1511.07289}, 2015.

\bibitem{kingma2015adam}
D.~P. Kingma and J.~Ba, ``Adam: A method for stochastic optimization,'' in
  \emph{Proceedings of the International Conference on Learning Representations
  (ICLR)}, 2015.

\end{thebibliography}
\label{sec:biography}

\begin{IEEEbiography}[{\includegraphics[width=1in,height=1.25in,clip,keepaspectratio]{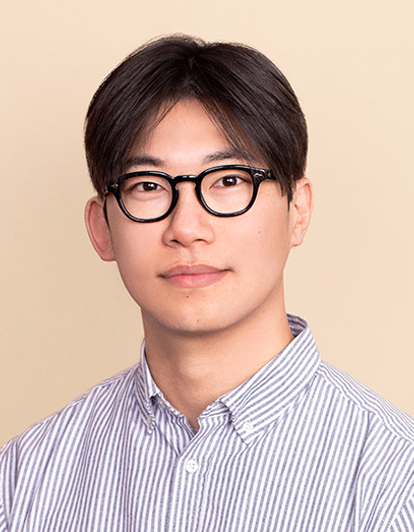}}]{Yunyong Ko} is an assistant professor in the School of Computer Science and Engineering, Chung-Ang University (CAU), Seoul, Korea.
He received the B.S. and Ph.D. degrees from Hanyang University in 2013 and 2021, respectively. 
Before joining CAU, he was a postdoctoral researcher in the Department of Computer Science, University of Illinois at Urbana-Champaign (UIUC); a postdoctoral researcher with Hanyang University, Seoul, Korea.
His research interests include large-scale data mining and machine learning across various data types, such as graphs/hypergraphs, texts, and images, for real-world applications.
\end{IEEEbiography}

\begin{IEEEbiography}[{\includegraphics[width=1in,height=1.25in,clip,keepaspectratio]{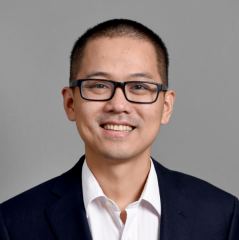}}]{Hanghang Tong}
is an associate professor in the Department of Computer Science, University of Illinois at Urbana-Champaign (UIUC). 
He received the B.S. degree in automation from Tsinghua University, in 2002, and the M.S. and Ph.D. degrees in machine learning from Carnegie Mellon University (CMU), in 2008 and 2009, respectively. 
Before joining UIUC, he was an assistant professor with Arizona State University; an assistant professor with the City University of New York; and a research staff member with IBM T. J. Watson Research Center. 
His research interests include large scale data mining for graphs and multimedia. He has received several awards, including Best Paper Award in CIKM 2012, SDM 2008, and ICDM 2006.
He is a fellow of the IEEE and a distinguished member of the ACM.
\end{IEEEbiography}

\begin{IEEEbiography}[{\includegraphics[width=1in,height=1.25in,clip,keepaspectratio]{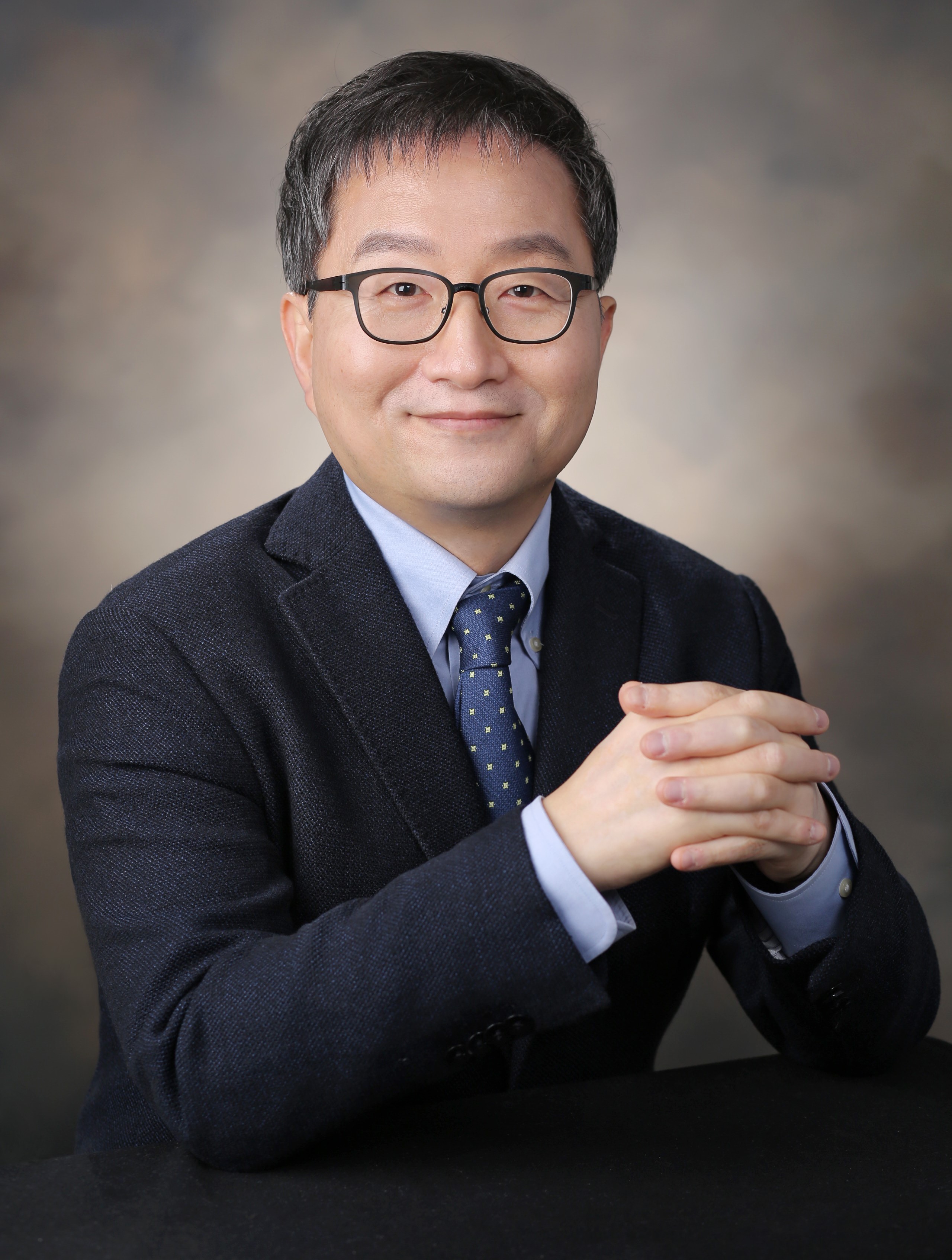}}]{Sang-Wook Kim} is a Distinguished Professor in the Department of Computer Science, Hanyang University, Seoul, Korea.
He received the B.S. degree in computer engineering from Seoul National University, in 1989, and the M.S. and Ph.D. degrees in computer science from the Korea Advanced Institute of Science and Technology (KAIST), in 1991 and 1994, respectively. From 1995 to 2003, he served as an associate professor with Kangwon National University. 
From 2009 to 2010, he visited the Computer Science Department, Carnegie Mellon University, as a visiting professor. From 1999 to 2000, he worked with the IBM T. J. Watson Research Center, USA, as a postdoc. He also visited the Computer Science Department of Stanford University as a visiting researcher in 1991. 
His research interests include databases, data mining, multimedia information retrieval, social network analysis, recommendation, and web data analysis. He is a member of the ACM and the IEEE.
\end{IEEEbiography}

\vfill

\end{document}